\documentclass{article}

\usepackage[nonatbib,preprint]{style_ips}

\usepackage[utf8]{inputenc} 
\usepackage[T1]{fontenc}    
\usepackage{hyperref}       
\usepackage{url}            
\usepackage{booktabs}       
\usepackage{amsfonts}       
\usepackage{nicefrac}       
\usepackage{microtype}      
\usepackage{xcolor}         
\usepackage{bm}
\usepackage{graphicx}
\usepackage{amsmath}
\usepackage{multirow}

\usepackage[linesnumbered,ruled]{algorithm2e}
\def\x{\bm{x}}
\def\f{\bm{f}}
\def\y{\bm{y}}

\def\w{\bm{w}}
\def\A{\bm{A}}

\def\s{\bm{s}}

\def\I{\bm{I}}
\def\F{\bm{F}}
\def\V{\bm{V}}

\def\C{\bm{C}}
\newcommand{\E}{\mathrm{E}}
\newtheorem{theorem}{Theorem}

\newtheorem{definition}{Definition}
\newtheorem{proposition}{Proposition}

\DeclareMathOperator*{\argmin}{arg\,min}

\title{Score-Based Variational Inference for Inverse Problems}

\author{%
  Zhipeng Xue \thanks{Zhipeng Xue and Penghao Cai are with the Pervasive Communication Center, Purple Mountain Laboratories, Nanjing 211111, China (e-mail: {xuezhipeng,caipenghao}@pmlabs.com.cn)}\\
  \And
  Penghao Cai $^{\ast}$\\
  \And
  Xiaojun Yuan \thanks{Xiaojun Yuan is with the National Key Laboratory of Wireless Communications, University of Electronic Science and Technology of China, Chengdu 611731, China (e-mail: xjyuan@uestc.edu.cn)}\\
  \And
  Xiqi Gao \thanks{ Xiqi Gao is with the National Mobile Communications Research Laboratory, Southeast University, Nanjing 210096, China, and also with Purple Mountain Laboratories, Nanjing 211111, China (e-mail: xqgao@seu.edu.cn)}\\
}

\begin{document}

\maketitle

\begin{abstract}
Existing diffusion-based methods for inverse problems sample from the posterior using score functions and accept the generated random samples as solutions. In applications that posterior mean is preferred, we have to generate multiple samples from the posterior which is time-consuming. In this work, by analyzing the probability density evolution of the conditional reverse diffusion process, we prove that the posterior mean can be achieved by tracking the mean of each reverse diffusion step. Based on that, we establish a framework termed reverse mean propagation (RMP) that targets the posterior mean directly. We show that RMP can be implemented by solving a variational inference problem, which can be further decomposed as minimizing a reverse KL divergence at each reverse step. We further develop an algorithm that optimizes the reverse KL divergence with natural gradient descent using score functions and propagates the mean at each reverse step. Experiments demonstrate the validity of the theory of our framework and show that our algorithm outperforms state-of-the-art algorithms on reconstruction performance with lower computational complexity in various inverse problems.

\end{abstract}

\section{Introduction}
Diffusion models \cite{sohl2015deep,song2019generative, ho2020denoising,song2020denoising,rombach2022high} have shown impressive performance for image generation. For diffusion models such as diffusion denoising probability model (DDPM) \cite{ho2020denoising} and denoising score matching with Langevin dynamics (SMLD) \cite{song2019generative}, the essential part is the learning of score functions of data distributions with large datasets. By approximating score functions with neural networks such as U-Net \cite{ronneberger2015u,song2019generative}, the prior of complex data distributions can be learned implicitly which encourages many applications. Inverse problems aim to recover an unknown state $\x_0$ from observation $\y$, which is fundamental to various research areas such as wireless communication, image processing and natural language processing. Recent works \cite{jalal2020robust, kawar2021snips, song2020denoising, song2021solving, chung2022improving, kawar2022denoising, meng2022diffusion, chung2022diffusion, laumont2022bayesian, mardani2023variational} have shown that diffusion models can be used for solving inverse problems since the prior of data distribution is learned implicitly with score functions. 

Based on Bayes' rule, diffusion models are used for the generation of data from the posterior distribution with score functions of data and likelihood, and thus can be applied in solving reverse problems. The main difficulty of applying diffusion models to solving inverse problems is the calculation of likelihood score. SNIPS \cite{kawar2021snips} and DDRM \cite{kawar2022denoising} are proposed to solve noisy linear inverse problems with diffusion process in the spectral domain. In these methods, the measurement and data to be estimated are transformed into the spectral domain via singular value decomposition (SVD), and the conditional score can be calculated with SVD explicitly. In \cite{meng2022diffusion}, the authors propose to approximate the likelihood score by a noise-perturbed pseudo-likelihood score which has a closed form under certain assumptions and can be efficiently calculated using SVD for noisy linear inverse problems. MCG \cite{chung2022improving} circumvents the calculation of likelihood score by projections onto the measurement constrained manifold. In DPS \cite{chung2022diffusion}, the Laplace method is used for the approximation of the likelihood score for general inverse problems. Instead of directly approximating the likelihood score, a variational inference based method termed RED-Diff \cite{mardani2023variational} optimizes the reconstruction loss with score matching regularization.

\begin{figure}
    \centering
    \includegraphics[width=0.9\textwidth]{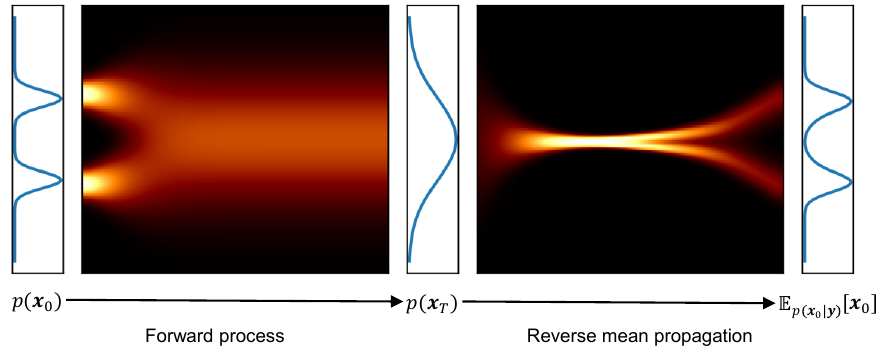}
    \caption{An illustration of Reverse Mean Propagation (RMP) for Gaussian mixture model. In the experiment, the data prior is $p(x_0) = \frac{1}{2}(\mathcal{N}(x_0;\mu_1,v_1^2)+ \mathcal{N}(x_0;\mu_2,v_2^2))$ and measurement $y = a x + v_0 \varepsilon$ where $\mu_1=-1$, $\mu_2=1$, $v_1=v_2=0.2$, $v_0=0.5$, $a=1$, $\varepsilon\sim \mathcal{N}(0,1)$ and $T=1000$. RMP is deterministic when $y$ and $\x_T$ are given and the final output converges to $\E_{p(x_0|y)}[x_0]$.}
    \label{fig:process_rmp}
\end{figure}

The works mentioned above mainly focus on generating random samples from the posterior distribution rather than finding a statistical estimate such as the posterior mean, i.e. the MMSE estimation in Bayesian inference, which is preferred in many applications \cite{kay1993fundamentals}. To get the posterior mean, we have to sample from the posterior multiple times and average out which is time-consuming. In this work, we show that the evolution of the conditional probability of the reverse diffusion process can be characterized by the reverse mean and covariance, and the posterior mean can be achieved by tracking the mean of each conditional reverse diffusion step. Base on this, we propose a variational inference based framework that minimizes the reverse KL divergence and propagate the reverse mean at each reverse step, referred to as Reverse Mean Propagation (RMP). By implementing RMP with stochastic natural gradient descent and approximating the gradient with score functions, we devise a practical algorithm for inverse problems. Instead of generating samples by running algorithms multiple times, RMP outputs an estimate which is closed to the posterior mean directly and thus can reduce complexity greatly. We demonstrate the validity of the theory of RMP by a toy example as illustrated in Fig \ref{fig:process_rmp} and show that by adopting the RMP algorithm, we achieve a large improvement over existing state-of-the-art algorithms on various image reconstruction tasks.

\section{Background}
\subsection{Inverse Problems}
An inverse problem is defined as the estimation of an unknown state or a latent $\x_0\in \mathbb{R}^{N\times 1}$ from measurement $\y\in \mathbb{R}^{M\times 1}$. Specifically, the measurement process can be described by a measurement operator $\mathcal{A}:\mathbb{R}^{N\times 1}\rightarrow \mathbb{R}^{M\times 1}$, and the final output is a noisy version of the measurement:
\begin{align}
    \y = \mathcal{A}(\x_0) + \w_0 \label{inv_prob}
\end{align}
where $\w_0 \in \mathbb{R}^{M\times 1}$ is the measurement noise. Usually the measurement noise is assumed to be zero mean Gaussian with variance $\epsilon^2\I$. Other noise models may also apply. In linear inverse problems, the measurement function $\mathcal{A}$ is linear and can be represented by a linear transform $\A \in \mathbb{R}^{M\times N}$. We focus on inverse problems with known $\mathcal{A}$ in this paper.

\subsection{Variational Inference}
The inverse problem (\ref{inv_prob}) can be formulated as a Bayesian estimation problem. The posterior distribution of $\x_0$ is given by $p(\x_0|\y) = \frac{p(\y|\x_0)p(\x_0)}{p(\y)}$,
where $p(\x_0)$ is the prior distribution of $\x_0$ and $p(\y|\x_0)$ is the conditional distribution of $\y$ given $\x_0$. The posterior mean, i.e., the MMSE estimator can be employed for the estimation of $\x_0$. However, the posterior $p(\x_0|\y)$ is intractable in general since the prior $p(\x_0)$ and the likelihood $p(\y|\x_0)$ may be very complicated in real applications. An alternative way is to find an approximation of the posterior as in variational inference (VI). VI introduces a distribution $q_{\phi}(\x_0|\y)$ and maximizes a lower bound of the log probability of marginal $p(\y)$:
 \begin{align}
 \begin{split}
     \log p(\y) & = \log \int \frac{q_{\phi}(\x_0|\y)}{q_{\phi}(\x_0|\y)} p(\y|\x_0) p(\x_0) d\x_0\\
 & \geq  \int q_{\phi}(\x_0|\y) \log \frac{p(\y,\x_0)  }{q_{\phi}(\x_0|\y)} d\x_0 = -\mathcal{F}_{\phi}(\y)\\
     & = - \text{KL}(q_{\phi}(\x_0|\y)||p(\x_0|\y)) + \log p(\y) \label{elbo}
 \end{split}
 \end{align}
where the inequality is obtained by using the Jensen’s inequality. The lower bound is referred as the evidence lower bound (ELBO) or the negative of free energy $\mathcal{F}(\y)$. It is worth noting that maximizing ELBO is equivalent to minimizing the KL divergence between $q_{\phi}(\x_0|\y)$ and $p(\x_0|\y)$ as shown in the last line of (\ref{elbo}). Many methods have been proposed for the optimization of (\ref{elbo}) such as mean field VI \cite{blei2017variational}, black box VI \cite{ranganath2014black}, stochastic VI \cite{kingma2013auto} and normalizing flow VI \cite{rezende2015variational}. However, these methods are difficult to be applied to real applications since prior $p(\x_0)$ is complicated and is usually learned by neural networks. It is shown that learning the distribution of high dimensional data through score matching \cite{vincent2011connection} directly is inaccurate since the existance of low density data regions \cite{song2019generative}. Also, perturbing data with Gaussian noise makes the data distribution more amenable to learn \cite{song2019generative} which is the core of score-based generative model.

\subsection{Score-Based Generative Models}
Score-based generative models or diffusion models generate samples of a data distribution from the reverse process of a diffusion process. The diffusion process is also called the forward process where Gaussian noise is added gradually to the original data distribution until the noisy data are approximately Gaussian-distributed. More specifically, the diffusion process is a Markov process with joint probability of its latent states $\{\x_{k}\}_{k=0}^{T}$ given by
\begin{align}
    p(\x_{T:0}) =p(\x_{0}) \prod_{k=0}^{T-1} p(\x_{k+1}|\x_{k})\label{diffusion_process}.
\end{align}
Two classes of widely studied diffusion models, i.e., the variance preserving (VP) diffusion model \cite{ho2020denoising} and the variance exploding (VE) diffusion model \cite{song2019generative} are distinguished by Markov diffusion kernel $p(\x_{k+1}|\x_{k})$. For VE diffusion, $p(\x_{k+1}|\x_{k}) = \mathcal{N}(\x_{k+1};\x_{k},(\sigma_{k+1}^2-\sigma_{k}^2)\I)$, where $\sigma_T^2>\sigma_{T-1}^2>\cdots > \sigma_1^2 >\sigma_0^2 = 0$, and for VP diffusion, $p(\x_{k+1}|\x_{k}) = \mathcal{N}(\x_{k+1};\sqrt{1-\beta_{k+1}}\x_{k},\beta_{k+1}\I)$ where $\beta_T>\beta_{T-1}>\cdots > \beta_1 >\beta_0 = 0$. The learning of score function $\nabla_{\x_{k}} \log p(\x_{k})$ is essential to the sample generation for both diffusion models, and thus such models are called score-based generative models. In VP diffusion models, a variational reverse process is leaned to minimize the KL divergence between forward and reverse process. The score functions of perturbed data distributions are trained with a variational bound. Samples are generated from the learned reverse process with ancestral sampling method. In VE diffusion models, the score functions of perturbed
data distributions are learned with a score network using denosing score matching \cite{vincent2011connection}. Samples from the desired distribution are generated with Langevin dynamics method.

\section{Diffusion Process and Posterior Estimation}\label{diffusion_estimation}
The measurement $\y$ in (\ref{inv_prob}) and diffusion states $\{\x_k\}_{k=0}^{T}$ in (\ref{diffusion_process}) form a new Markov chain $\y\rightarrow\x_0 \rightarrow \x_{1}\cdots \rightarrow\x_T$ and the reverse conditional $p_k(\x_k|\x_{k+1},\y)$ is given by $p_{k}(\x_k|\x_{k+1},\y)  = \frac{p(\x_{k+1}|\x_{k})p(\x_k|\y)}{p(\x_{k+1}|\y)}$, $\forall k=0,\cdots T-1$. In this part, we focus on the property of reverse conditional $p_k(\x_k|\x_{k+1},\y)$. We relate the Markov chain $\{\x_k\}_{k=0}^{T}$ to continuous stochastic process $\{\x_t\}_{t=0}^{1}$ by letting $\x_k = \x_{t=k \Delta t}$, where $\Delta t = \frac{1}{T}$. Then, the discrete diffusion process (\ref{diffusion_process}) becomes a continuous process in the limit $\Delta t \rightarrow 0$. We have the following results in the limit of $\Delta t \rightarrow 0$.

\begin{proposition}\label{prop2}
    For diffusion models with forward process (\ref{diffusion_process}), the reverse conditional $p_{k}(\x_k|\x_{k+1},\y)$, $\forall k = 0\cdots T-1$, is Gaussian when $\Delta t \rightarrow 0$. For VE and VP diffusion, the mean and covariance of $p_{k}(\x_k|\x_{k+1},\y)$ are tractable with mean given by
\begin{align}
    \begin{split}
     \bm{\mu}_{k}(\x_{k+1},\y) &= \V_{k,1} \x_{k+1} + \V_{k,2}\E_{p(\x_{0}|\y)}[\x_{0}]
\end{split}\label{reverse_mean}
\end{align}
where $\V_{k,1} =(\sigma_{k}^2\I + \C_{\x_{0}}) (\sigma_{k+1}^2\I+\C_{\x_{0}})^{-1}$ and $\V_{k,2}=(\sigma_{k+1}^2-\sigma_{k}^2)(\sigma_{k+1}^2\I+\C_{\x_{0}})^{-1}$ for VE diffusion, and $\V_{k,1} = \sqrt{\alpha_{k+1}}((1-\bar{\alpha}_{k})\I + \bar{\alpha}_{k}\C_{\x_{0}})((1-\bar{\alpha}_{k+1})\I + \bar{\alpha}_{k+1}\C_{\x_{0}})^{-1} $ and $\V_{k,2}=\sqrt{\bar{\alpha}_{k}}(1-\alpha_{k+1} )((1-\bar{\alpha}_{k+1})\I+ \bar{\alpha}_{k+1}\C_{\x_{0}})^{-1}$ for VP diffusion. $\E_{p(\x_{0}|\y)}[\x_{0}]$ and $\C_{\x_{0}}$ are the mean and covariance of $p(\x_0|\y)$ respectively. $\bar{\alpha}_k = \prod_{i=0}^{k}\alpha_i$, $\alpha_i=1-\beta_i$. The covariance of $p_{k}(\x_k|\x_{k+1},\y)$ for VE and VP diffusion models are given respectively by
\begin{align}
    \begin{split}
        \C_{k,\text{VE}} &= (\sigma_{k+1}^2-\sigma_{k}^2)(\sigma_{k}^2\I + \C_{\x_{0}}) (\sigma_{k+1}^2\I+\C_{\x_{0}})^{-1}\\
        \C_{k,\text{VP}} &= \frac{\beta_{k+1}}{1-\beta_{k+1}}((1-\bar{\alpha}_{k})\I+\bar{\alpha}_{k}\C_{\x_{0}}) \left(\left(\frac{\beta_{k+1}}{1-\beta_{k+1}}+1-\bar{\alpha}_{k}\right)\I+\bar{\alpha}_{k}\C_{\x_{0}}\right)^{-1}.
    \end{split}
\end{align}
\end{proposition}
Proposition \ref{prop2} generalizes the Gaussian property of $p_{t}(\x_t|\x_{t+\Delta t})$ to the conditional case $p_{t}(\x_t|\x_{t+\Delta t},\y)$ when $\Delta t \rightarrow 0$. The essential is that the reverse diffusion process can also be expressed by a reverse SDE \cite{song2020score}. Based on Proposition \ref{prop2}, we obtain the following main result.

\begin{definition}
The reverse mean propagation chain of a diffusion process is defined as
\begin{align}
     \bm{\mu}_{T} \rightarrow \bm{\mu}_{T-1}(\x_{T}=\bm{\mu}_{T},\y) \rightarrow \cdots \rightarrow  \bm{\mu}_{1}(\x_{2}= \bm{\mu}_{2},\y) \rightarrow   \bm{\mu}_{0}(\x_{1}= \bm{\mu}_{1},\y)\label{reverse_chain}
\end{align}
where $\bm{\mu}_{k}(\x_{k+1}=\bm{\mu}_{k+1},\y)$ is the mean of $p_{k}(\x_k|\x_{k+1} = \bm{\mu}_{k+1},\y)$, $\forall k=0,\cdots,T-1$, and $\bm{\mu}_{T}$ and $\bm{\mu}_{0}$ are the initial point and the end point of the reverse chain respectively.
\end{definition}

\begin{theorem}\label{theorem1}
For VE diffusion, when $\Delta t \rightarrow 0$, the end point of the reverse chain, i.e., $\bm{\mu}_{0}$ is given by
 \begin{align}
    \bm{\mu}_{0}   = (\sigma_{0}^2\I + \C_{\x_{0}}) (\sigma_{T}^2\I+\C_{\x_{0}})^{-1} \bm{\mu}_{T}+ 
 (\sigma_{T}^2-\sigma_{0}^2)(\sigma_{T}^2\I+\C_{\x_{0}})^{-1}\E_{p(\x_{0}|\y)}[\x_{0}]
\end{align}
and $\bm{\mu}_{0}  \rightarrow \E_{p(\x_{0}|\y)}[\x_{0}]$ as $\sigma_T\rightarrow \infty$. For VP diffusion, when $\Delta t \rightarrow 0$, $\bm{\mu}_{0}$ is given by
\begin{align}
\begin{split}
   \bm{\mu}_{0}  =\ & \sqrt{\bar{\alpha}_{T}}((1-\bar{\alpha}_{0})\I+\bar{\alpha}_{0}\C_{\x_{0}})((1-\bar{\alpha}_{T})\I+\bar{\alpha}_{T}\C_{\x_{0}})^{-1} \bm{\mu}_{T} \\
    &+ (1-\bar{\alpha}_{T}) ((1-\bar{\alpha}_{T})\I+\bar{\alpha}_{T}\C_{\x_{0}})^{-1}\E_{p(\x_{0}|\y)}[\x_{0}]
\end{split}
\end{align}
and $\bm{\mu}_{0}  \rightarrow \E_{p(\x_{0}|\y)}[\x_{0}]$ as $\bar{\alpha}_{T}\rightarrow 0$.
\end{theorem}
According to Theorem \ref{theorem1}, the posterior mean can be obtained by tracking the mean at each reverse step. By calculating the reverse mean and following the reverse chain in (\ref{reverse_chain}), we get a posterior estimation framework termed Reverse Mean Propagation (RMP) for inverse problems as presented in Algorithm \ref{RMP}. We note that when the initial point of the reverse chain $\bm{\mu}_{T}$ and $\y$ are given, the reverse chain is deterministic and converges to the posterior mean $\E_{p(\x_{0}|\y)}[\x_{0}]$.

\begin{algorithm}
    \SetKwInOut{Input}{Input}
    \SetKwInOut{Output}{Output}

    \Input{$\y$, $T$, $\bm{\mu}_T$}
    \For{$k = T-1 : 0$}{
    Propagate the reverse mean: $\x_{k+1}=\bm{\mu}_{k+1}$\\ 
    Calculate the reverse mean of $p_{k}$: $\bm{\mu}_{k}(\x_{k+1} = \bm{\mu}_{k+1},\y) = \E_{p_k(\x_{k}|\x_{k+1}=\bm{\mu}_{k+1},\y)}[\x_k]$\\ 
    }
    \Output{$\bm{\mu}_{0}$}
    \caption{Reverse Mean Propagation (RMP)}\label{RMP}
\end{algorithm}

\section{Score-Based Variational Inference}
In this section, we propose a score-based variational inference method to implement the RMP framework. We show that tracking the mean of the reverse process of each step can be formulated as a sequential of variational inference problems which we prove to be equivalent to a variational inference problem for all latent variables. We solve the variational inference by stochastic natural gradient descent with approximations that simplify the calculation.
\subsection{RMP as Variational Inference}
In Section \ref{diffusion_estimation}, we present the RMP framework based on the reverse diffusion process. However, in practice, the reverse mean in RMP cannot be calculated using (\ref{reverse_mean}) since $\E_{p(\x_0|\y)}[\x_0]$ and $\C_{\x_0}$ are unknown. We now show that the RMP framework can be implemented using variational inference. Instead of applying variational inference on the conditional posterior of $\x_{0}$ as in (\ref{elbo}), we focus on the joint conditional posterior of $\{\x_{k}\}_{k=0}^{T}$, i.e., $p(\x_{0:T}|\y)$, which includes all the latent variables in the diffusion process. The variational reverse process with joint conditional is defined by
\begin{align}
    q(\x_{0:T}|\y) = q(\x_T|\y)\prod_{k=0}^{T-1} q_{k}(\x_{k}|\x_{k+1},\y),
\end{align}
where $q_{k}(\x_{k}|\x_{k+1},\y) = \mathcal{N}(\x_{k};\bm{\mu}_{k}(\x_{k+1},\y),\C_{k}(\x_{k+1},\y)), \forall k=0:T-1$. We set $q(\x_{T}|\y) = \mathcal{N}(\x_{T};0,\I)$. The variational reverse process is chosen as a Markov chain since the reverse of the diffusion process (\ref{diffusion_process}) is a Markov process. The KL divergence between variational joint posterior $q(\x_{0:T}|\y)$ and and joint posterior $p(\x_{0:T}|\y)$ is given by
\begin{align}
    KL(q||p) =\int q(\x_{0:T}|\y) \log \frac{q(\x_{0:T}|\y)}{p(\x_{0:T}|\y)}d \x_{0:T}\label{kl_qp}
\end{align}
where the forward joint posterior $p(\x_{0:T}|\y) = p(\x_T|\y)\prod_{k=0}^{T-1} p(\x_{k}|\x_{k+1:T},\y)$. For VE diffusion $p(\x_{T}|\y) = \mathcal{N}(\x_{T};0,\sigma_T^2\I)$ and for VP diffusion $p(\x_{T}|\y) = \mathcal{N}(\x_{T};0,\I)$. The following proposition simplifies the minimization of $KL(q||p)$ with proof given in Appendix \ref{proof_prop1}.

\begin{proposition}\label{prop_1}
For a diffusion process with forward process (\ref{diffusion_process}), the KL divergence between variational $q(\x_{0:T}|\y)$ and joint posterior $p(\x_{0:T}|\y)$ defined in (\ref{kl_qp}) equals
\begin{align}
    KL(q||p) = \sum_{k=T-1}^{0}\int q(\x_{k+1}|\y)\int q_{k}(\x_{k}|\x_{k+1},\y) \log \frac{q_k(\x_k|\x_{k+1},\y)}{p_{k}(\x_k|\x_{k+1},\y)} d\x_{k} d \x_{k+1},
\end{align}
and the minimization of $KL(q||p)$ is equivalent to the minimization of
\begin{align}
    KL(q_{k}||p_{k}) = \int q_{k}(\x_{k}|\x_{k+1},\y) \log \frac{q_k(\x_k|\x_{k+1},\y)}{p_{k}(\x_k|\x_{k+1},\y)} d\x_{k}, \forall k=0,\cdots,T-1.\label{kl_qk_pk}
\end{align}
\end{proposition}
According to Proposition \ref{prop_1}, we can minimize the KL divergence between $q$ and $p$ by minimizing the KL divergence $KL(q_k||p_k)$, i.e., solve the following VI problem at each reverse step $k$:
\begin{align}
     q_k^{\star} = \argmin_{q_k} KL( q_k||p_k), \forall k=0,\cdots,T-1.\label{mini_qkpk}
\end{align}
By propagating the mean of $q_k$ and solving problem (\ref{mini_qkpk}) at each reverse step $k$, $\forall k=0,\cdots,T-1$, we can approximate the RMP framework in Algorithm \ref{RMP} based on variational inference, as detailed below.

\subsection{Variational Inference by Natural Gradient Descent}
For $q_{k}(\x_{k}|\x_{k+1},\y) = \mathcal{N}(\x_{k};\bm{\mu}_{k},\Lambda_k^{-1}\I)$, the KL divergence between $q_k$ and $p_{k}$ is given by
\begin{align}
\begin{split}
   KL(q_{k}||p_{k}) &=\int q_{k}(\x_{k}|\x_{k+1},\y) \log \frac{q_k(\x_k|\x_{k+1},\y)}{p_{k}(\x_k|\x_{k+1},\y)} d\x_{k}\\
    &= - \frac{N}{2} \log (2\pi/\Lambda_{k})  - \frac{N}{2} -\E_{ q_{k}}[\log p_{k}(\x_k|\x_{k+1},\y)].\label{kl_qjpj}
\end{split}
\end{align}
A common practice to optimize $KL(q_{k}||p_{k})$ is to update variational parameters $\phi_{k} = \{\bm{\mu}_{k},\Lambda_{k}\}$ using mini-batch stochastic gradient descent which involves the calculation of $\nabla_{\phi_k} KL(q_{k}||p_{k})$. Since $q_k$ is Gaussian, the gradient of $\E_{ q_{k}}[\log p_{k}(\x_k|\x_{k+1},\y)]$ in (\ref{kl_qjpj}) with respect to variational parameters $\phi_{k} = \{\bm{\mu}_{k},\Lambda_{k}\}$ have simple forms \cite{opper2009variational} which are given by:
\begin{align}
\begin{split}
    \nabla_{\bm{\mu}_k} \E_{q_k}[\log p_{k}(\x_k|\x_{k+1},\y)]  &= \E_{q_k}[\nabla_{\x_k}\log p_{k}(\x_k|\x_{k+1},\y)]\label{grad_mu}\\
    \nabla_{\bm{\Lambda}_k} \E_{q_k}[\log p_{k}(\x_k|\x_{k+1},\y)] &= -\frac{1}{2} \Lambda_k^{-2}\E_{q_k} [\text{Tr}(\nabla^2_{\x_k} \log p_{k}(\x_k|\x_{k+1},\y)]
\end{split}
\end{align}
where $\nabla_{\x_k}\log p_{k}(\x_k|\x_{k+1},\y)$ and $\nabla^2_{\x_k} \log p_{k}(\x_k|\x_{k+1},\y)$ are the gradient and Hessian of $\log p_{k}(\x_k|\x_{k+1},\y)$ respectively, and $\text{Tr}(\cdot)$ returns the trace of the input matrix.

As a special case of steepest descent, gradient descent updates parameter that lies in the Euclidean space. However, our objective is to optimize parameters that represent a distribution, it makes sense to take the steepest descent direction in the distribution space. As in natural gradient descent \cite{martens2020new}, the parameter to be optimized lies on a Riemannian manifold and we choose the steepest descent direction along that manifold. Thus, we choose KL-divergence as the metric of distribution space and take steepest descent in this space. For KL-divergence metric, the natural gradient of parameter of a loss function $\mathcal{L} = \E_{q_{\phi}}[h(\x)]$ is defined as $\tilde{\nabla}_{\phi}\mathcal{L} =  \F_{\phi}^{-1} \nabla_{\phi}\mathcal{L}$ \cite{martens2020new}, where $\F_{\phi}$ is the Fisher information matrix of $\phi$ given by the variance of the gradient of log probability of parameter, i.e., $\text{Cov}_{q_{\phi}}[\nabla_{\phi}\log q_{\phi}(\x)]$. For $q(\x) = \mathcal{N}(\x;\bm{\mu},\bm{\Sigma}) = \mathcal{N}(\x;\bm{\mu}, \Lambda^{-1}\I)$, the Fisher information matrices of mean and precision are given respectively by $\F_{\bm{\mu}} = \Lambda\I$ and $\F_{\Lambda} = \frac{1}{2}  \Lambda^{-2}\I$. Thus, the natural gradients of parameters $\{\bm{\mu}, \Lambda\}$ have concise forms given by
\begin{align}
\begin{split}
    \tilde{\nabla}_{\bm{\mu}}\mathcal{L} &=\F_{\bm{\mu}}^{-1} \nabla_{\bm{\mu}} \E_{q}[h(\x)] = \Lambda^{-1}\E_{q}[\nabla_{\x}h(\x)]\\
    \tilde{\nabla}_{\Lambda}\mathcal{L}&=\F_{\Lambda}^{-1} \nabla_{\Lambda} \E_{q}[h(\x)] = -\E_{q} [\text{Tr}(\nabla^2_{\x} h(\x))].
\end{split}\label{natural_grad}
\end{align}

Following the natural gradient given in (\ref{natural_grad}), we update the variational parameters $\phi_{k} = \{\bm{\mu}_{k},\Lambda_{k} = v_{k}^{-1}\}$ of loss function in (\ref{kl_qjpj}) using natural gradient descent (NGD) as 
\begin{align}
\begin{split}
    \bm{\mu}_{k} &\leftarrow \bm{\mu}_{k} - s_1 \Lambda_{k}^{-1} \nabla_{\mu_k}KL(q_{k}||p_{k}) =\bm{\mu}_{k} + s_1 \Lambda_{k}^{-1} \E_{ q_{k}} [\nabla_{\x_{k}} \log p_{k}(\x_k|\x_{k+1},\y) ] \\
    \Lambda_{k} &\leftarrow \Lambda_{k} - 2 s_2  \Lambda_{k}^{2} \nabla_{\Lambda_k}KL(q_{k}||p_{k}) = \Lambda_{k} - s_2 (N \Lambda_{k} + \E_{ q_{k}} [\text{Tr}(\nabla^2_{\x_{k}} \log p_{k}(\x_k|\x_{k+1},\y)) ])\label{ngd_update}
\end{split}
\end{align}
where $s_1$ and $s_2$ are step sizes. We obtain a stochastic NGD update when the expectations of gradient and Hessian matrix in (\ref{ngd_update}) are approximated by sample mean:
\begin{align}
\begin{split}
    \bm{\mu}_{k}^{(i+1)} & =  \bm{\mu}_{k}^{(i)} + s_1 (\Lambda_{k}^{(i)})^{-1} \frac{1}{L}\sum_{i=1}^{L}\nabla_{\x_{k}} \log p_{k}(\x_k|\x_{k+1},\y)  |_{\x_k = \x_k^{(i)}\sim q_k^{(i)}}\\
    \Lambda_{k}^{(i+1)} & =   \Lambda_{k}^{(i)} - s_2 \left(N \Lambda_{k}^{(i)} + \frac{1}{L}\sum_{i=1}^{L}\text{Tr}(\nabla^2_{\x_{k}} \log p_{k}(\x_k|\x_{k+1},\y)) |_{\x_k = \x_k^{(i)} \sim q_k^{(i)}}\right)
\end{split}\label{sngd_update}
\end{align}
where $L$ is the number of samples. The stochastic update of parameters of $q_k$ converges to the local minima of KL divergence $KL(q_{k}||p_{k})$ which is a Gaussian approximation of the posterior $p_{k}(\x_k|\x_{k+1},\y)$. It is worth noting that we choose stochastic NGD since it achieves a good performance and parameters involved are easy to tune in our experiments. Other optimization methods may be applied. In stochastic NGD update (\ref{sngd_update}), the gradient and Hessian of $\log p_{k}(\x_k|\x_{k+1},\y)$ are required. We next introduce some approximations to simplify the calculation.

\subsection{Score-Based Gradient Calculation}
From Bayes' rule, the score of reverse conditional $p(\x_k|\x_{k+1},\y)  = \frac{p(\x_{k+1}|\x_{k})p(\x_k|\y)}{p(\x_{k+1}|\y)}$ involved in (\ref{sngd_update}) is given by $\nabla_{\x_k} \log p_{k}(\x_k|\x_{k+1},\y) = \nabla_{\x_k} \log p(\x_{k+1}|\x_{k}) + \nabla_{\x_k} \log p(\y|\x_k) + \nabla_{\x_k} \log p(\x_k) $, where $\nabla_{\x_k} \log p(\x_k)$ is the noisy score function which can be approximated by a well-trained score network $\s_{\bm{\theta}}(\x_k,\sigma_k)$ and $\nabla_{\x_k} \log p(\x_{k+1}|\x_{k})$ can be calculated explicitly for both VE and VP diffusion models. For VE diffusion $\nabla_{\x_k} \log p(\x_{k+1}|\x_{k}) = \frac{\x_{k+1}-\x_{k}}{\sigma_{k+1}^2 - \sigma_k^2}$, and for VP diffusion $\nabla_{\x_k} \log p(\x_{k+1}|\x_{k}) = \frac{\sqrt{1-\beta_{k+1}}}{\beta_{k+1}} \x_{k+1} - \frac{1-\beta_{k+1}}{\beta_{k+1}}\x_k$. However, the likelihood score, i.e., the gradient of logarithm conditional $\nabla_{\x_k} \log p(\y|\x_k)$ is hard to handle in general. For linear inverse problems, several SVD based approximations of $\nabla_{\x_k} \log p(\y|\x_k)$ are proposed in \cite{kawar2021snips,kawar2022denoising,meng2022diffusion} for linear measurements and Gaussian approximation for general measurements are discussed in \cite{song2023pseudoinverse}. In \cite{chung2022diffusion}, the authors propose the following approximation that can be applied for general measurements:
\begin{align}
    \log p(\y|\x_k) \approx \log p(\y|\hat{\x}_0(\x_{k}))\label{likelihood_score_approx}
\end{align}
where $\hat{\x}_0(\x_{k})$ is the MMSE estimate of $\x_0$. For VE diffusion, according to the Tweedie formula:
\begin{align}
    \hat{\x}_0(\x_{k}) = \E_{p(\x_{0}|\x_{k})}[\x_0]=\x_{k} + \sigma_k^2 \nabla_{\x_k} \log p(\x_k).\label{tweedie_1}
\end{align}
Similarly, for VP diffusion, the MMSE estimation of $\x_0$ given $\x_{k}$ is
\begin{align}
    \hat{\x}_0(\x_{k}) = \E_{p(\x_{0}|\x_{k})}[\x_0]=\frac{1}{\sqrt{\bar{\alpha}_k}}(\x_{k} + (1-\bar{\alpha}_k) \nabla_{\x_k} \log p(\x_k)).\label{tweedie_2}
\end{align}
The approximation error of (\ref{likelihood_score_approx}) can be quantified with Jenson's Gap as given in \cite{chung2022diffusion}. We choose the likelihood approximation of (\ref{likelihood_score_approx}) in our implementation, other approximation methods can be applied to RMP as discussed in Appendix. As a conclusion, the gradient is calculated as
\begin{align}
     \nabla_{\x_k} \log p_{k}(\x_k|\x_{k+1},\y)\approx \nabla_{\x_k} \log p(\x_{k+1}|\x_{k})+ \gamma_k \nabla_{\x_k} \log p(\y|\hat{\x}_0(\x_{k})) + \s_{\bm{\theta}}(\x_k,\sigma_k).
\end{align}
where the parameter $\gamma_k$ is added to balance the approximated likelihood score and prior score. We set $\gamma_k = \zeta   \frac{\|\s_{\bm{\theta}}(\x_k,\sigma_k)\|_2 }{\|\nabla\log p(\y|\hat{\x}_0(\x_{k}))\|_2}$ where $\zeta$ is a hyper parameter to be tuned for different problems. The idea behind the strategy is that we should always keep a balance between the data score and the likelihood score.

\begin{algorithm}
    \SetKwInOut{Input}{Input}
    \SetKwInOut{Output}{Output}
    \Input{$\y$, $s_1$, $T$, $T_{in}$, $T_{s}$, $\x_{T}$, $\bm{\mu}_{T-1}^{0}$}
    \For{$k = T-1 : 0$}{
    For VE $\Lambda_{k}^{-1} = \frac{\sigma_k^2(\sigma_{k+1}^2-\sigma_k^2)}{\sigma_{k+1}^2}$if $k > T_{s}$ else $\Lambda_{k}^{-1} = \sigma_{k+1}^2-\sigma_k^2$ (for VP $\Lambda_{k}^{-1} = \beta_{k+1}$)\\
    \For{$i = 0:T_{in}-1$}{
    $\bm{\mu}_{k}^{(i+1)} = \bm{\mu}_{k}^{(i)} + s_1 \Lambda_{k}^{-1} \left(\nabla_{\x_k} \log p(\x_{k+1}|\x_{k}) + \gamma_k \nabla_{\x_k}  \log p(\y|\hat{\x}_0(\x_{k})) + \s_{\bm{\theta}}(\x_k,\sigma_k)\right)$\\
    where $\x_k\sim \mathcal{N}(\x;\bm{\mu}_{k}^{(i)}, \Lambda_{k}^{-1}\I)$, $\hat{\x}_0(\x_{k}) \! = \! \x_{k} + \sigma_k^2 \s_{\bm{\theta}}(\x_k,\sigma_k)$ for VE \\
    (for VP $\hat{\x}_0(\x_{k}) \! = \! \frac{1}{\sqrt{\bar{\alpha}_k}}(\x_{k} + (1\!-\!\bar{\alpha}_k) \s_{\bm{\theta}}(\x_k,\sigma_k)$)
    }
     $\x_{k} = \bm{\mu}_{k}^{(T_{in})}$ and $\bm{\mu}_{k-1}^{(0)} = \bm{\mu}_{k}^{(T_{in})}$
    }
    \Output{$\bm{\mu}_{0}^{(T_{in})}$}
    \caption{VE/VP-RMP with Score-based Stochastic NGD}\label{RMP2}
\end{algorithm}

\subsection{Fixed Precision Update}

In the update (\ref{sngd_update}), the Hessian matrix $\nabla^2_{\x_{k}} \log p_{k}(\x_k|\x_{k+1},\y)$ is difficult to acquire in general. Also, the complexity involved in the calculation of Hessian may prevent the application of the algorithm. Thus, we introduce an approximation that does not require the calculation of Hessian. According to the Proposition \ref{prop2}, the update of precision $\Lambda_{k}^{(i+1)}$ converges to the precision of $p_{k}(\x_k|\x_{k+1},\y)$. Thus, we can fix the update of $\Lambda_{k}^{(i+1)}$ in Algorithm \ref{RMP} to the precision of $p_{k}(\x_k|\x_{k+1},\y)$ and only update $\bm{\mu}_{k}$ at each step. According to Proposition \ref{prop2}, for VE diffusion model, if we set $\C_{\x_0} = v_{\x_0}\I$, then the inverse of precision is given by $(\Lambda_{k}^{(i)})^{-1}  = \frac{(\sigma_k^2+v_{\x_{0}})(\sigma_{k+1}^2-\sigma_k^2)}{\sigma_{k+1}^2 + v_{\x_{0}}}$. We cannot calculate $(\Lambda_{k}^{(i)})^{-1}$ directly since $v_{\x_{0}}$ is unknown. However, for the cases that $k$ is close to $T$, $\sigma_k^2$ is large enough comparing to $v_{\x_{0}}$. Thus, $(\Lambda_{k}^{(i)})^{-1}$ can be approximated by $\frac{\sigma_k^2(\sigma_{k+1}^2-\sigma_k^2)}{\sigma_{k+1}^2}$. For the case that $k$ is close to $0$, $\sigma_{k+1}^2, \sigma_k^2\rightarrow 0$, we have $(\Lambda_{k}^{(i)})^{-1} \approx \sigma_{k+1}^2-\sigma_k^2$. Similarly, for VP diffusion model, if we set $\C_{\x_0} = v_{\x_0}\I$, then $(\Lambda_{k}^{(i)})^{-1} =\frac{\frac{\beta_{k+1}}{1-\beta_{k+1}}(1-\bar{\alpha}_{k}+\bar{\alpha}_{k}v_{\x_{0}})}{\frac{\beta_{k+1}}{1-\beta_{k+1}}+(1-\bar{\alpha}_{k}+\bar{\alpha}_{k}v_{\x_{0}})}$. Since $\beta_{k}\approx 0$ and $\bar{\alpha}_{k}\approx 0$, we have $(\Lambda_{k}^{(i)})^{-1} \approx  \beta_{k+1} $. With stochastic NGD based VI and score-based approximations of gradient and Hessian, we summarize an algorithm given in Algorithm \ref{RMP2}.

\section{Experiments}

\begin{figure}[!ht]
    \centering
    \includegraphics[width=0.8\textwidth]{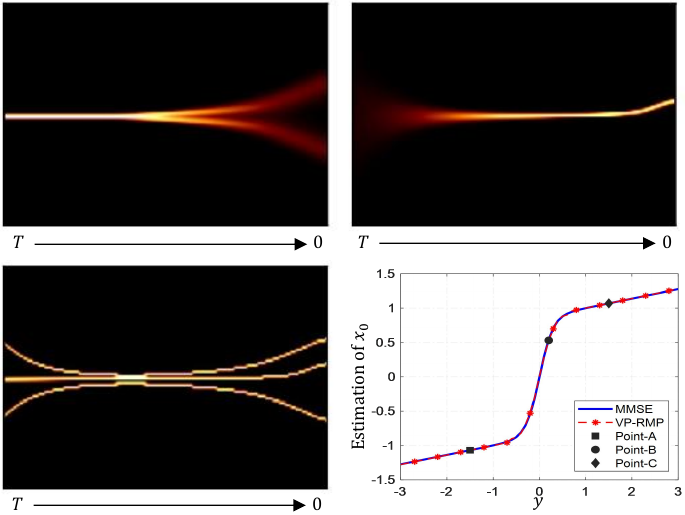}
    \caption{Illustration of the process of VP-based RMP for Gaussian mixture model. Top-left: Various measurement $y$ with fixed $x_T = 0$. Top-right: Random $x_T$ with $y=0.2$. Bottom-left: $(x_T, y)= (-1,-1.5)$, $(x_T, y)= (0,0.2)$ and $(x_T, y)= (1,1.5)$. Bottom-right: MMSE estimation $\E_{p(x_0|y = \bar{y})}[x_0]$ and VP-RMP outputs for different measurement $y$.}
    \label{fig:rmp_comp}
\end{figure}
\subsection{Gaussian Mixture Model}
We generate $x_0$ from a Gaussian mixture prior $p(x_0) = \frac{1}{2}\mathcal{N}(x_0;\mu_1,v_1^2)+\frac{1}{2} \mathcal{N}(x_0;\mu_2,v_2^2)$ and get measurement from $y = a x_0 + v_0 \varepsilon$, where $\varepsilon\sim \mathcal{N}(0,1)$, to demonstrate the rationale of RMP. In the experiments, we set $\mu_1=1$, $\mu_2=-1$, $v_1=v_2=0.2$, $v_0=0.5$, $a=1$, and $T=1000$. Since the model is simple, the data and likelihood score can be calculated exactly (given in Appendix \ref{gaussian_mixture}). By running VP-RMP in Algorithm \ref{RMP2} (with exact score), the propagation of the reverse mean is shown in Figs. \ref{fig:process_rmp} and \ref{fig:rmp_comp}. In the right part of Fig. \ref{fig:process_rmp}, $x_{T}$ is initialized randomly from $p_{T}(x_T)$ for various measurement $y$. We see from the plot that the final output of VP-RMP, i.e., the reverse mean $\mu_{0}$ converges to $\E_{p(x_0|y)}[x_0]$. In the top-left plot of Fig. \ref{fig:rmp_comp}, for various measurement $y$, we initialize $x_T=0$. The final output of VP-RMP is almost the same as that in Fig. \ref{fig:process_rmp} which converges to $\E_{p(x_0|y)}[x_0]$. In the top-right plot of Fig. \ref{fig:rmp_comp}, $x_{T}$ is initialized randomly from $p_{T}(x_T)$ and $y$ is set to 0.2. The reverse mean converges to the posterior mean $\E_{p(x_0|y=0.2)}[x_0]$ no matter what $x_T$ is. In the bottom-left plot of Fig. \ref{fig:rmp_comp}, $x_{T}$ and $y $ are fixed to three cases. We see that the evolution of reverse mean is almost deterministic when the initial point $x_T$ and $y$ are fixed. In the bottom-right plot of Fig. \ref{fig:rmp_comp}, the theoretical results of $\E_{p(x_0|y = \bar{y})}[x_0]$ and estimations of $x_0$ using VP-RMP for various measurements $y$ are shown where points A, B, and C correspond to the outputs of VP-RMP for the three cases given in the bottom-left plot respectively. We see from the plot that the outputs of VP-RMP consists with the posterior mean well.

\begin{table}[!ht]
\centering
\caption{Quantitative results (PSNR, SSIM, FID, LPIPS) of solving linear inverse problems with Gaussian noise ($\epsilon=0.05$) on FFHQ validation dataset. $\textbf{Bold}$: best, $\underline{\text{Underline}}$: second best.}
\footnotesize
\setlength\tabcolsep{2pt}
\renewcommand{\arraystretch}{1.2}
\begin{tabular}{ccccccccccc}
\hline
\multirow{2}{*}{Methods}       & \multicolumn{2}{c}{SR (4 $\times$)}                       & \multicolumn{2}{c}{Inpaint (box)}               & \multicolumn{2}{l}{Inpaint (random)} & \multicolumn{2}{c}{Deblur (Gauss) }                & \multicolumn{2}{c}{Deblur (motion) }                \\ \cline{2-11} 
     & \multicolumn{1}{c}{PSNR $\uparrow$} & SSIM $\uparrow$         & \multicolumn{1}{c}{PSNR $\uparrow$} & SSIM $\uparrow$                 & \multicolumn{1}{l}{PSNR $\uparrow$}    & SSIM $\uparrow$  & \multicolumn{1}{c}{PSNR $\uparrow$} & SSIM $\uparrow$                 & \multicolumn{1}{c}{PSNR $\uparrow$} & SSIM $\uparrow$                 \\ \hline
VE-RMP (Ours)      & $\bm{29.26}$    & $\bm{0.8421}$             &  $25.23$   &  ${0.8361}$                     &    $\bm{34.58}$     &  $\underline{0.9315}$       &   $\bm{28.28}$   &   $\bm{0.8214}$        &   $\underline{27.74}$    &    $\bm{0.8232}$                   \\ 
VP-RMP (Ours)& $\underline{28.89}$     & $\underline{0.8410}$ &  $\bm{25.78}$   & $\bm{0.8716}$  &   $\underline{34.27}$     &  $\bm{0.9291}$       & $\underline{27.96}$    & $\underline{0.8153}$ &   $\bm{28.10}$  & \underline{0.7921} \\  
DPS         & \multicolumn{1}{c}{24.32}     &  0.7016       & 24.68    &                0.8182       & 29.85     &  0.8554       & 25.50    &  0.7210     &   23.42   &  ${0.6712}$                      \\ 
MCG                           & 18.16    &       0.2145                & 11.30     &   0.2450                    & 11.06        &   0.0812      & 11.65    &  0.1317         & 11.53    & 0.1065                      \\ 
\hline
\multirow{2}{*}{Methods}       & \multicolumn{2}{c}{SR (4 $\times$)}                       & \multicolumn{2}{c}{Inpaint (box)}               & \multicolumn{2}{l}{Inpaint (random)} & \multicolumn{2}{c}{Deblur(Gauss) }                & \multicolumn{2}{c}{Deblur (motion) }                \\ \cline{2-11} 
     & \multicolumn{1}{c}{FID $\downarrow$} & LPIPS $\downarrow$         & \multicolumn{1}{c}{FID $\downarrow$} & LPIPS $\downarrow$                 & \multicolumn{1}{l}{FID $\downarrow$}    & LPIPS $\downarrow$  & \multicolumn{1}{c}{FID $\downarrow$} & LPIPS $\downarrow$                 & \multicolumn{1}{c}{FID $\downarrow$} & LPIPS $\downarrow$                 \\ \hline
VE-RMP (Ours)  & 89.04    &   0.2278    &   96.36   &   0.2372                   &    ${36.37}$     &   $\underline{0.0948}$      &   94.43   &     0.2515     &  90.63    & $\underline{0.2438}$                      \\ 
VP-RMP (Ours)   & $\bm{57.27}$     &   $\bm{0.1890}$    &   $\bm{28.36}$    &  $\bm{0.1250}$    &   $\bm{22.41}$     &  $\bm{0.0771}$       &  $\bm{59.10}$   & $\bm{0.2024}$ &  
$\bm{62.80}$    & $\bm{0.2383}$ \\ 
DPS    &${72.44 }$     &    ${0.2484}$  & ${53.35 }$     &  ${0.1905 }$                      & 57.74       &    0.1887     & $\underline{60.18 }$    & ${0.2377 }$           & ${68.83}$      &   ${0.2576 }$                    \\ 
MCG                            & 227.65     &         0.6232              & 443.24     &  0.7929                     & 486.09       &   0.8224      & 354.01    &  0.7760                     & 461.38  &       0.7222                \\ 
\hline
\end{tabular}\label{table_psnr_ssim}
\end{table}

\subsection{Image Reconstruction}
To show the advantage of RMP over posterior sampling based algorithms, we compare our proposed VE/VP-RMP in Algorithm \ref{RMP2} with baseline algorithms for different image reconstruction tasks on FFHQ $256\times 256$ dataset \cite{karras2019style}. For VP-based methods, the pre-trained score network for FFHQ $256\times 256$ is take from \cite{chung2022diffusion} and for VE-based methods, the pre-trained score networks for FFHQ $256\times 256$ is taken from \cite{song2020score}. The algorithms we compare include Diffusion Posterior Sampling (DPS) \cite{chung2022diffusion}, Manifold Constrained Gradients (MCG). Algorithms such as Denoising Diffusion Restoration Models (DDRM) and $\Pi$GDM \cite{song2023pseudoinverse} are discussed in the appendix part since these methods have difference likelihood score approximation and DDRM can only be applied for linear measurements. Unless otherwise specified, the measurement noise is set to Gaussian. The following metrics are used for comparisons: Frechet Inception Distance (FID) \cite{heusel2017gans}, Learned Perceptual Image Patch Similarity (LPIPS) \cite{zhang2018unreasonable}, Peak Signal-to-Noise-Ratio (PSNR) and Structural Similarity Index Measure (SSIM) \cite{wang2004image}.

For VP-based algorithms i.e., DPS, MCG and VP-RMP, $\beta_k$ varies linearly with $k$ in range from 0.0001 to 0.02 as given in \cite{ho2020denoising}. The timesteps for DPS, MCG, and VP-RMP are all fixed to $T=400$. The inner loop size $T_{in} = 1$ for VP-RMP. For VE-RMP, we set $T=30$ and $T_{in}=20$ for all tasks. $\{\sigma_k\}, k=1,2\cdots, T$ is set as a positive geometric sequence as suggested in \cite{song2019generative} that satisfies $\frac{\sigma_{k}}{\sigma_{k+1}} >1$. In all experiments, we set $\sigma_k = \sigma_{min}(\frac{\sigma_{max}}{\sigma_{min}})^{\frac{k-1}{T-1}}$ where $\sigma_{min}=0.01$ and $\sigma_{max} = 100$ that matches the pre-trained score network given in \cite{song2020score}. The settings of step size $s_1$ and $\zeta$ for VE/VP-RMP are given in Appendix \ref{params_setting}. Other parameters in DPS, MCG are set according to the configurations provided by its' authors. The code of our algorithms is developed under the framework developed by DPS, and the measurement operator used in our comparisons are the same for all algorithms such that the comparisons are fair enough for all algorithms. Our code is available at \url{https://github.com/neuripsrmp/rmp.git}.

\textbf{Linear Inverse Problems:} We compare algorithms on linear inverse problems of image reconstruction including super resolution, inpainting (box, random) and deblur (Gaussian deblur, motion deblur). The evaluation results on different metrics are given in Tables \ref{table_psnr_ssim}. We see from Tables \ref{table_psnr_ssim} that VE/VP-RMP outperforms other algorithms on almost all tasks and VP-RMP achieves the best performance in both FID and LPIPS metrics in all tasks. We present some results in Fig. \ref{fig:all_results}. More results on FFHQ and ImageNet \cite{russakovsky2015imagenet} datasets can be found in Appendix \ref{results}.

\textbf{Nonlinear Inverse Problems:} We compare VE/VP-based RMP with DPS on non-linear image reconstruction tasks including phase retrieval and non-linear deblur. The comparison results is presented in Table \ref{table_nonlinear}. We see that similarly to the linear cases, RMP achieves better performance comparing to DPS in most of the metrics.

\begin{figure}[!ht]
    \centering
    \includegraphics[width=0.95\textwidth]{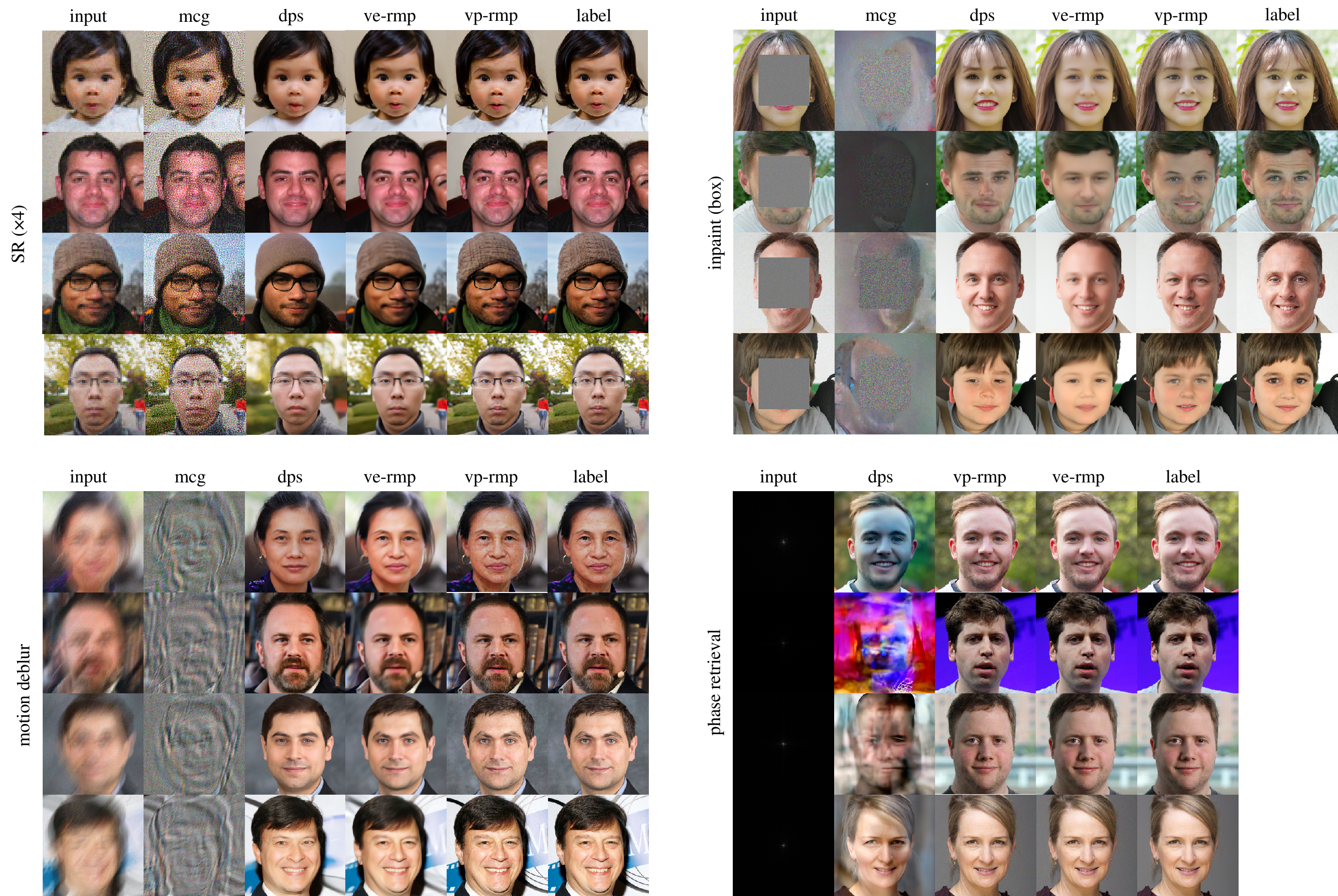}
    \caption{Part of the results on solving inverse problems with Gaussian noise ($\epsilon=0.05$).}
    \label{fig:all_results}
\end{figure}

\begin{table}[!ht]
\centering
\caption{Quantitative evaluation (PSNR, SSIM, FID, LPIPS) of solving nonlinear inverse problems on FFHQ validation dataset. $\textbf{Bold}$: best, $\underline{\text{Underline}}$: second best.}
\footnotesize
\setlength\tabcolsep{5pt}
\begin{tabular}{ccccccccc}
\hline
\multirow{2}{*}{Methods}       & \multicolumn{4}{c}{Phase retrieval}                                                     & \multicolumn{4}{c}{Nonlinear deblur}                                \\ \cline{2-9} 
     & \multicolumn{1}{c}{PSNR} & \multicolumn{1}{c}{SSIM} & \multicolumn{1}{l}{FID} & LPIPS & \multicolumn{1}{c}{PSNR} & \multicolumn{1}{c}{SSIM} & {FID} & LPIPS \\ \hline
VE-RMP (Ours)                      & $\bm{27.98}$     & $\bm{0.7860}$      & $\underline{72.84}$     & $\underline{0.1857}$      &  $\bm{ 25.21}$    & $\bm{0.7222}$     &  143.02   &  
 ${0.3401}$     \\  
\multicolumn{1}{c}{VP-RMP (Ours)} &  $\underline{25.32} $    & $\underline{0.7632}$      &  $\bm{62.11}$    &  $\bm{0.1512}$     & $\underline{24.33}$    &  $\underline{0.6872}$     & $\bm{66.47}$   &   $\bm{0.2219}$    \\  
DPS         & 12.75    & ${0.4142 }$    & ${218.78}$    &   0.5828    & 23.62  &  ${0.6696}$    & $\underline{76.85}$    & $\underline{0.2685}$     \\ \hline
\end{tabular}
\label{table_nonlinear}
\end{table}

\begin{figure}[!ht]
    \centering
    \includegraphics[width=0.47\textwidth]{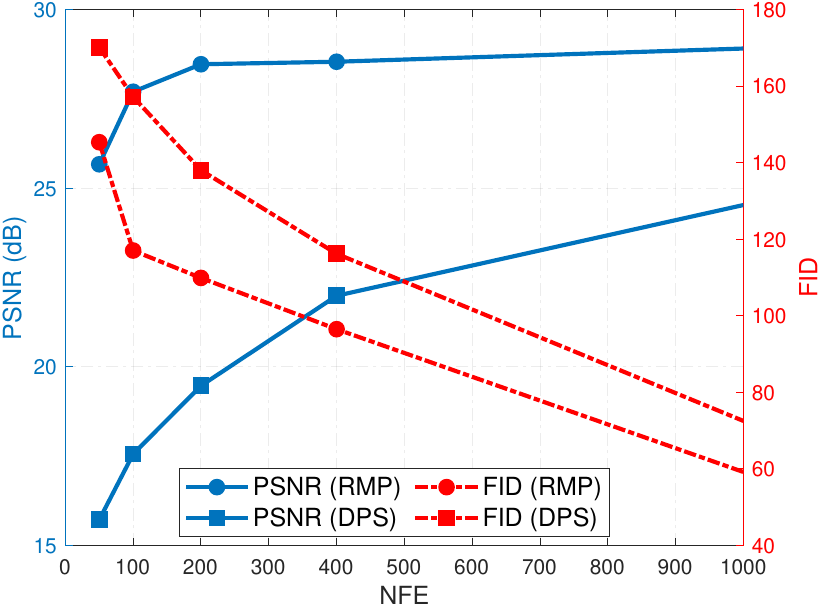}
    \includegraphics[width=0.48\textwidth]{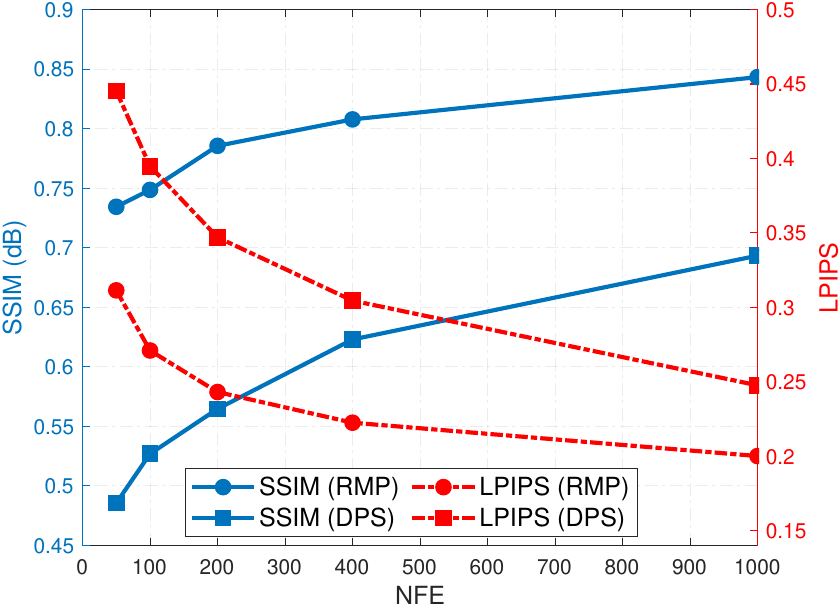}
    \caption{Performance comparisons of algorithms with different NFE on SR task.}
    \label{fig:nfe_performance}
\end{figure}

\subsection{Complexity vs Performance}
The complexity of diffusion-based methods dependent on the number of neural function evaluations (NFE). For our algorithm, RMP consists of $T$ outer loops and each outer loop has $T_{in}$ inner loops. The operations in the outer loops can be ignored since only variable assignments and scalar calculation are involved. In each inner loop, two main operations are involved: denoising with score network $\s_{\bm{\theta}}(\x_k,\sigma_k)$ and stochastic natural gradient descent step which requires the likelihood score. Thus, the NFE of VE/VP-RMP is $T T_{in}$. To show the performance of RMPs with different NFE, we compare the curves of NFE versus the performance of VP-RMP and DPS on SR task in Fig. \ref{fig:nfe_performance}. From the figure, we see that RMP outperforms DPS significantly on all metrics for all NFEs. We also note that for the PSNR metric, VP-RMP with $50$ NFEs achieves a higher PSNR than DPS with $1000$ NFEs.

\section{Colculsions}
In this paper, we propose a score-based variational inference framework for inverse problems. By optimizing the reverse KL divergence at each step of the reverse process and tracking the evolution of reverse mean at each reverse step, a practical algorithm for general inverse problems is proposed. Extensive experiments validate our theoretical results and demonstrate that our proposed algorithm achieves superior performance on various image reconstruction tasks.


\appendix


\section{Proof of Proposition \ref{prop2}}\label{proof_prop2}
\subsection{Gaussian Property}
First, the reverse conditional $p_{k}(\x_k|\x_{k+1},\y)$ is given by
\begin{align}
   p_{k}(\x_k|\x_{k+1},\y)  = \frac{p(\x_{k+1}|\x_{k})p(\x_k|\y)}{p(\x_{k+1}|\y)}
\end{align}
where $\x_{k}$ and $\x_{k+1}$ can be viewed as the discretization of continuous diffusion process of $\x_t$. We relate $\{\x_k\}_{k=0}^{T}$ to continuous stochastic process $\{\x_t\}_{t=0}^{1}$ by letting $\x_k = \x_{t=k \Delta t}$, where $\Delta t = \frac{1}{T}$. The discrete $\{\x_k\}_{k=0}^{T}$ becomes a continuous process in the limit $\Delta t \rightarrow 0$. Thus, we focus on the continuous form of the reverse conditional, i.e., $p_{k}(\x_k|\x_{k+1},\y) = p(\x_{t}|\x_{t+\Delta t},\y)$. The reverse conditional in a continuous form is given by
\begin{align}
    p(\x_{t}|\x_{t+\Delta t },\y) = \frac{p(\x_{t+\Delta t}|\x_{t})p(\x_t|\y)}{p(\x_{t+\Delta t}|\y)}.
\end{align}
As shown in \cite{song2020score}, the forward process can be characterized by the following SDE:
\begin{align}
    d\x = \f_t(\x)dt +g_t(\x) d\bar{\w}
\end{align}
where $\bar{\w}$ is the standard Wiener process. For VE diffusion, $\f_t(\x) = 0$ and $g_t(\x) = \sqrt{\frac{d\sigma(t)^2}{d t}}$, and for VP diffusion, $\f_t(\x) = -\frac{1}{2}\beta(t)\x$ and $g_t(\x) = \sqrt{\beta(t)}$. When $\Delta t \rightarrow 0$ we have
\begin{align}
    \x_{t+\Delta t} - \x_{t} = \f_t(\x_t)\Delta t + g_t(\x_t) \sqrt{\Delta t}  \bm{\varepsilon}
\end{align}
where $\bm{\varepsilon}\sim \mathcal{N}(0,\I)$. Then
\begin{align}
\begin{split}
    p(\x_{t+\Delta t}|\x_{t})& = \mathcal{N}(\x_{t+\Delta t};\x_{t}+\f_t(\x_t)\Delta t,g_t^2(\x_t)   \Delta t)\\
    &\propto \exp{\left(-\frac{\|\x_{t+\Delta t}-\x_{t}-\f_t(\x_t)\Delta t\|_2^2}{2 g_t^2(\x_t)   \Delta t}\right)}
\end{split}
\end{align}
and
\begin{align}
    p(\x_{t}|\x_{t+\Delta t},\y) \propto \exp{\left( -\frac{\|\x_{t+\Delta t}-\x_{t}-\f_t(\x_t)\Delta t\|_2^2}{2 g_t^2(\x_t)   \Delta t} \! +\!\log p(\x_t|\y)\! -\! \log p(\x_{t+\Delta t}|\y)\right)}.
\end{align}
When $\Delta t\rightarrow 0$, $\log p(\x_{t}|\y)$ can be expressed by Taylor expansion:
 \begin{align}
     \log p(\x_{t+\Delta t}|\y) \approx \log p(\x_t|\y) + (\x_{t+\Delta t}-\x_t)^T\nabla_{\x_t} \log p(\x_t|\y) + \Delta t \frac{\partial}{\partial t}\log p(\x_t|\y).
 \end{align}
Thus, we have 
 \begin{align}
 \begin{split}
     p(\x_{t}|\x_{t+\Delta t},\y)  
     \propto    \exp{\left(-\frac{\|\x_{t+\Delta t}-\x_t -(\f_t(\x_t)-g_t^2(\x_t)\nabla_{\x_t}\log p(\x_t|\y))\Delta t\|_2^2}{2g_t^2(\x_t) \Delta t}+\mathcal{O}(\Delta t)\right)}
 \end{split}\label{pxt_xt1y}
 \end{align}
which is Gaussian when $\Delta t \rightarrow 0$. Thus, $p_{k}(\x_{k}|\x_{k+1},\y)$ is Gaussian when $\Delta t \rightarrow 0$.

\subsection{VE Diffusion}

For VE diffusion models, the mean of $p(\x_k|\y)$ is given by
\begin{align}
\begin{split}
    \E_{p(\x_k|\y)}[\x_{k}]& = \int \x_{k}p(\x_k|\y)d\x_{k}\\
 &= \int \x_{k}\int p(\x_k|\x_{0})p(\x_{0}|\y)d\x_{0} d\x_{k}\\
 &= \int \int \x_{k} p(\x_k|\x_{0}) d\x_{k}p(\x_{0}|\y)d\x_{0}\\
 &=\int \x_{0}p(\x_{0}|\y) d\x_{0} = \E_{p(\x_{0}|\y)}[\x_{0}]
\end{split}\label{mu_xk_y}
\end{align}
and its covariance matrix is given by
\begin{align}
\begin{split}
&Cov_{p(\x_k|\y)}[\x_{k}]\\
 =\ & \int \x_{k}\x_{k}^Tp(\x_k|\y)d\x_{k} - \E_{p(\x_{0}|\y)}[\x_{0}] \E_{p(\x_{0}|\y)}[\x_{0}]^T\\
 =\ & \int \x_{k}\x_{k}^T\int p(\x_k|\x_{0})p(\x_{0}|\y)d\x_{0} d\x_{k}- \E_{p(\x_{0}|\y)}[\x_{0}] \E_{p(\x_{0}|\y)}[\x_{0}]^T\\
 = \ &\int \int \x_{k}\x_{k}^T p(\x_k|\x_{0}) d\x_{k}p(\x_{0}|\y)d\x_{0}- \E_{p(\x_{0}|\y)}[\x_{0}] \E_{p(\x_{0}|\y)}[\x_{0}]^T\\
 =\ &\int (\x_{0}\x_{0}^T+\sigma_{k}^2\I) p(\x_{0}|\y) d\x_{0} - \E_{p(\x_{0}|\y)}[\x_{0}] \E_{p(\x_{0}|\y)}[\x_{0}]^T\\
 =\ &\sigma_{k}^2\I + Cov_{p(\x_{0}|\y)}[\x_{0}]+ \E_{p(\x_{0}|\y)}[\x_{0}] \E_{p(\x_{0}|\y)}[\x_{0}]^T - \E_{p(\x_{0}|\y)}[\x_{0}] \E_{p(\x_{0}|\y)}[\x_{0}]^T\\
 =\ &\sigma_{k}^2\I + Cov_{p(\x_{0}|\y)}[\x_{0}] =\sigma_{k}^2\I + \C_{\x_{0}}.
\end{split}\label{c_xk_y}
\end{align}

Since $p_{k}(\x_{k}|\x_{k+1},\y)$ can be expanded as a Gaussian around $\x_{k+1}$ using the second order Taylor approximation when $\Delta t\rightarrow 0$ and $p(\x_{k+1}|\x_{k}) = \mathcal{N}(\x_{k+1};\x_{k},(\sigma_{k+1}^2-\sigma_{k}^2)\I)$, then $p(\x_{k}|\y)$ can also be expanded around $\x_{k+1}$ and approximately expressed in the form of a Gaussian when $\Delta t\rightarrow 0$, i.e., $p(\x_{k}|\y) \approx \mathcal{N}(\x_{k};\E_{p(\x_{0}|\y)}[\x_{0}], (\sigma_{k}^2 \I+ \C_{\x_{0}}))$. For two multivariate Gaussian distribution $G_1(\x) = \mathcal{N}(\x,\bm{\mu}_1,\bm{\Sigma}_1)$ and $G_2(\x) = \mathcal{N}(\x,\bm{\mu}_2,\bm{\Sigma}_2)$, the product $G_1(\x)G_2(\x)$ is also Gaussian with mean and covariance given by
\begin{align}
\begin{split}
    \bm{\mu}_{3} &= \bm{\Sigma}_2(\bm{\Sigma}_1+\bm{\Sigma}_2)^{-1}\bm{\mu}_1 + \bm{\Sigma}_1(\bm{\Sigma}_1+\bm{\Sigma}_2)^{-1}\bm{\mu}_2\\
    \bm{\Sigma}_3 &= \bm{\Sigma}_1(\bm{\Sigma}_1 +\bm{\Sigma}_2 )^{-1}\bm{\Sigma}_2.
\end{split}\label{gaussian_mul}
\end{align}

Thus, the mean and covariance of $p_{k}(\x_k|\x_{k+1},\y)$ can be calculated according to (\ref{gaussian_mul}) and are given respectively by:
\begin{align}
\begin{split}
    \bm{\mu}_{k}  &=(\sigma_{k}^2\I + \C_{\x_{0}}) (\sigma_{k+1}^2\I +\C_{\x_{0}})^{-1} \x_{k+1}+(\sigma_{k+1}^2-\sigma_{k}^2)(\sigma_{k+1}^2\I +\C_{\x_{0}})^{-1}\E_{p(\x_{0}|\y)}[\x_{0}]\\
    &=\V_{k,p_1}\x_{k+1} + \V_{k,p_2}\E_{p(\x_{0}|\y)}[\x_{0}]\\
    \C_{k} &= (\sigma_{k+1}^2-\sigma_{k}^2)(\sigma_{k}^2\I + \C_{\x_{0}}) (\sigma_{k+1}^2\I+\C_{\x_{0}})^{-1}.
\end{split}\label{discrete_mu_c2}
\end{align}

\subsection{VP Diffusion}
For VP diffusion models, the mean of $p(\x_k|\y)$ is given by
\begin{align}
\begin{split}
    \E_{p(\x_k|\y)}[\x_{k}]& = \int \x_{k}p(\x_k|\y)d\x_{k}\\
 &= \int \x_{k}\int p(\x_k|\x_{0})p(\x_{0}|\y)d\x_{0} d\x_{k}\\
 &= \int \int \x_{k} p(\x_k|\x_{0}) d\x_{k}p(\x_{0}|\y)d\x_{0}\\
 &=\int \sqrt{\bar{\alpha}_{k}}\x_{0}p(\x_{0}|\y) d\x_{0} = \sqrt{\bar{\alpha}_{k}}\E_{p(\x_{0}|\y)}[\x_{0}]
\end{split}
\end{align}
and its covariance matrix is given by
\begin{align}
\begin{split}
    Cov_{p(\x_k|\y)}[\x_{k}]& = \int \x_{k}\x_{k}^Tp(\x_k|\y)d\x_{k} - \bar{\alpha}_{k}\E_{p(\x_{0}|\y)}[\x_{0}] \E_{p(\x_{0}|\y)}[\x_{0}]^T\\
 &= \int \x_{k}\x_{k}^T\int p(\x_k|\x_{0})p(\x_{0}|\y)d\x_{0} d\x_{k}- \bar{\alpha}_{k}\E_{p(\x_{0}|\y)}[\x_{0}] \E_{p(\x_{0}|\y)}[\x_{0}]^T\\
 &= \int \int \x_{k}\x_{k}^T p(\x_k|\x_{0}) d\x_{k}p(\x_{0}|\y)d\x_{0}- \bar{\alpha}_{k}\E_{p(\x_{0}|\y)}[\x_{0}] \E_{p(\x_{0}|\y)}[\x_{0}]^T\\
 &=\int (\bar{\alpha}_{k}\x_{0}\x_{0}^T+(1-\bar{\alpha}_{k})\I) p(\x_{0}|\y) d\x_{0} - \bar{\alpha}_{k}\E_{p(\x_{0}|\y)}[\x_{0}] \E_{p(\x_{0}|\y)}[\x_{0}]^T\\
 &=(1-\bar{\alpha}_{k})\I + \bar{\alpha}_{k}(Cov_{p(\x_{0}|\y)}[\x_{0}]+ \E_{p(\x_{0}|\y)}[\x_{0}] \E_{p(\x_{0}|\y)}[\x_{0}]^T)\\
 &\ \ \ \ - \bar{\alpha}_{k}\E_{p(\x_{0}|\y)}[\x_{0}] \E_{p(\x_{0}|\y)}[\x_{0}]^T\\
 &=(1-\bar{\alpha}_{k})\I + \bar{\alpha}_{k}Cov_{p(\x_{0}|\y)}[\x_{0}] =(1-\bar{\alpha}_{k})\I + \bar{\alpha}_{k}\C_{\x_{0}}
\end{split}
\end{align}
Similarly to the case of VE diffusion, since $p(\x_{k+1}|\x_{k}) = \mathcal{N}(\x_{k+1};\sqrt{1-\beta_{k+1}}\x_{k},\beta_{k+1}\I)$, the mean and covariance of $p_{k}(\x_k|\x_{k+1},\y)$ are given respectively by:
\begin{align}
\begin{split}
    \bm{\mu}_{k}  &= ((1-\bar{\alpha}_{k})\I + \bar{\alpha}_{k}\C_{\x_{0}})\left(\left(\frac{\beta_{k+1}}{1-\beta_{k+1}}+1-\bar{\alpha}_{k}\right)\I + \bar{\alpha}_{k}\C_{\x_{0}}\right)^{-1}\frac{\x_{k+1}}{\sqrt{1-\beta_{k+1}}} \\
    &\ \ \ \  +\frac{\beta_{k+1}}{1-\beta_{k+1}}\left(\left(\frac{\beta_{k+1}}{1-\beta_{k+1}}+1-\bar{\alpha}_{k}\right)\I + \bar{\alpha}_{k}\C_{\x_{0}}\right)^{-1}\sqrt{\bar{\alpha}_{k}}\E_{p(\x_{0}|\y)}[\x_{0}]\\
    &=((1-\bar{\alpha}_{k})\I + \bar{\alpha}_{k}\C_{\x_{0}})\left((1-\bar{\alpha}_{k+1})\I + \bar{\alpha}_{k+1}\C_{\x_{0}}\right)^{-1} \sqrt{\alpha_{k+1}}\x_{k+1} \\
    &\ \ \ \ +(1-\alpha_{k+1} )\left((1-\bar{\alpha}_{k+1})\I + \bar{\alpha}_{k+1}\C_{\x_{0}}\right)^{-1} \sqrt{\bar{\alpha}_{k}}\E_{p(\x_{0}|\y)}[\x_{0}]\\
    &=\V_{k,p_3}\x_{k+1} + \V_{k,p_4}\E_{p(\x_{0}|\y)}[\x_{0}]
\end{split}
\end{align}
and
\begin{align}
    \C_{k} =  \frac{\beta_{k+1}}{1-\beta_{k+1}}((1-\bar{\alpha}_{k})\I+\bar{\alpha}_{k}\C_{\x_{0}}) \left(\left(\frac{\beta_{k+1}}{1-\beta_{k+1}}+1-\bar{\alpha}_{k}\right)\I+\bar{\alpha}_{k}\C_{\x_{0}}\right)^{-1}.
\end{align}

\section{Proof of Theorem \ref{theorem1}}\label{proof_theorem1}

\subsection{VE Diffusion}
For VE diffusion model, if we set $\x_{k} = \bm{\mu}_{k}(\x_{k+1} = \bm{\mu}_{k+1},\y)$, $\forall k=0:T-1$, then from Proposition \ref{prop2}, we have
\begin{align}
\begin{split}
    &\bm{\mu}_{0}(\x_{1},\y)\\
    =\ & \V_{0,p_1}\x_{1} + \V_{0,p_2}\E_{p(\x_{0}|\y)}[\x_{0}]\\
    =\ &\V_{0,p_1}( \V_{1,p_1}\x_{2} + \V_{1,p_2}\E_{p(\x_{0}|\y)}[\x_{0}])+ \V_{0,p_2}\E_{p(\x_{0}|\y)}[\x_{0}]\\
    =\ &\V_{0,p_1}\V_{1,p_1}\x_{2} + \V_{0,p_1}\V_{1,p_2}\E_{p(\x_{0}|\y)}[\x_{0}] + \V_{0,p_2}\E_{p(\x_{0}|\y)}[\x_{0}]\\
    =\ & \cdots\\
    =\ & \prod_{i=0}^{T-1} \V_{i,p_1} \bm{\mu}_{T}+ (\V_{0,p_1}(\V_{1,p_1}(\cdots \V_{T-2,p_1}(\V_{T-1,p_2})+\V_{T-2,p_2})+\V_{1,p_2})+\V_{0,p_2})\E_{p(\x_{0}|\y)}[\x_{0}]).
\end{split}
\end{align}
For the first part, the coefficient of $\bm{\mu}_{T}$ equals
\begin{align}
    \prod_{i=0}^{T-1} \V_{i,p_1} = \prod_{i=0}^{T-1}(\sigma_{i}^2\I + \C_{\x_{0}}) (\sigma_{i+1}^2\I+\C_{\x_{0}})^{-1} = (\sigma_{0}^2\I + \C_{\x_{0}}) (\sigma_{T}^2\I+\C_{\x_{0}})^{-1}.
\end{align}
For the second part, $\forall k = 1:T-1$, we have 
\begin{align}
\begin{split}
    \V_{k-1,p_1}\V_{k,p_2}+\V_{k-1,p_2} &= (\sigma_{k-1}^2\I + \C_{\x_{0}}) (\sigma_{k}^2\I+\C_{\x_{0}})^{-1} (\sigma_{k+1}^2-\sigma_{k}^2)(\sigma_{k+1}^2\I+\C_{\x_{0}})^{-1} \notag\\
    &\ \ \ \ +(\sigma_{k}^2-\sigma_{k-1}^2)(\sigma_{k }^2\I+\C_{\x_{0}})^{-1}\\
    &= (\sigma_{k+1}^2-\sigma_{k-1}^2)(\sigma_{k+1}^2\I+\C_{\x_{0}})^{-1}.
\end{split}
\end{align}
Similarly, we get
\begin{align}
    (\V_{0,p_1}(\V_{1,p_1}(\cdots \V_{T-2,p_1}(\V_{T-1,p_2})+\V_{T-2,p_2})+\V_{1,p_2})+\V_{0,p_2}) = (\sigma_{T}^2-\sigma_{0}^2)(\sigma_{T}^2\I+\C_{\x_{0}})^{-1}
\end{align}
Thus, we have 
\begin{align}
    \bm{\mu}_{0}  = (\sigma_{0}^2\I + \C_{\x_{0}}) (\sigma_{T}^2\I+\C_{\x_{0}})^{-1} \bm{\mu}_{T}+ 
 (\sigma_{T}^2-\sigma_{0}^2)(\sigma_{T}^2\I+\C_{\x_{0}})^{-1}\E_{p(\x_{0}|\y)}[\x_{0}].
\end{align}
When $\sigma_T^2\rightarrow \infty$, $\bm{\mu}_{0}(\x_{1},\y) \rightarrow \E_{p(\x_{0}|\y)}[\x_{0}]$, which is the posterior mean.

\subsection{VP Diffusion}
Similarly to VE diffusion, for VP diffusion model, 
\begin{align}
\begin{split}
    &\bm{\mu}_{0}(\x_{1},\y) \\
    =\ & \V_{0,p_3}\x_{1} + \V_{0,p_4}\E_{p(\x_{0}|\y)}[\x_{0}]\\
    =\ & \prod_{i=0}^{T-1} \V_{i,p_3} \bm{\mu}_{T}+ (\V_{0,p_3}(\V_{1,p_3}(\cdots \V_{T-2,p_3}(\V_{T-1,p_4})+\V_{T-2,p_4})+\V_{1,p_4})+\V_{0,p_4})\E_{p(\x_{0}|\y)}[\x_{0}])
\end{split}
\end{align}

For the first part, the coefficient of $\bm{\mu}_{T}$ equals
\begin{align}
\begin{split}
    \prod_{i=0}^{T-1} \V_{i,p_3} &= \prod_{i=0}^{T-1}((1-\bar{\alpha}_{i})\I + \bar{\alpha}_{i}\C_{\x_{0}})\left((1-\bar{\alpha}_{i+1})\I + \bar{\alpha}_{i+1}\C_{\x_{0}}\right)^{-1} \sqrt{\alpha_{i+1}} \\
    &= \sqrt{\bar{\alpha}_{T}}((1-\bar{\alpha}_{0})\I+\bar{\alpha}_{0}\C_{\x_{0}})((1-\bar{\alpha}_{T})\I+\bar{\alpha}_{T}\C_{\x_{0}})^{-1}.
\end{split}
\end{align}
For the second part, $\forall i= 1:T-1$, we have 
\begin{align}
\begin{split}
    \V_{i-1,p_3}\V_{i,p_4}+\V_{i-1,p_4} &= ((1-\bar{\alpha}_{i-1})\I + \bar{\alpha}_{i-1}\C_{\x_{0}})\left(1-\bar{\alpha}_{i} + \bar{\alpha}_{i}\C_{\x_{0}}\right)^{-1} \\
     &\ \ \ \ \cdot \sqrt{\alpha_{i}}  (1-\alpha_{i+1} )\left((1-\bar{\alpha}_{i+1} )\I+ \bar{\alpha}_{i+1}\C_{\x_{0}}\right)^{-1} \sqrt{\bar{\alpha}_{i}}\\
 &\ \ \ \  + (1-\alpha_{i} )\left((1-\bar{\alpha}_{i})\I + \bar{\alpha}_{i}\C_{\x_{0}}\right)^{-1} \sqrt{\bar{\alpha}_{i-1}}\\
    &= (1-\alpha_i \alpha_{i+1})\sqrt{\bar{\alpha}_{i-1}} ((1-\bar{\alpha}_{i+1})\I+\bar{\alpha}_{i+1}\C_{\x_{0}})^{-1}.
\end{split}
\end{align}

Similarly, we obtain
\begin{align}
\begin{split}
    &\V_{0,p_3}(\V_{1,p_3}(\cdots \V_{T-2,p_3}(\V_{T-1,p_4})+\V_{T-2,p_4})+\V_{1,p_4})+\V_{0,p_4} \\
    =\ & (1-\alpha_{0}\alpha_1\cdots \alpha_{T})\sqrt{\bar{\alpha}_{0}}((1-\bar{\alpha}_{T})\I+\bar{\alpha}_{T}\C_{\x_{0}})^{-1}\\
    =\ &(1-\bar{\alpha}_{T}) ((1-\bar{\alpha}_{T})\I+\bar{\alpha}_{T}\C_{\x_{0}})^{-1}.
\end{split}
\end{align}
Thus, we have 
\begin{align}
\begin{split}
    \bm{\mu}_{0}(\x_{1},\y) =\ &\sqrt{\bar{\alpha}_{T}}((1-\bar{\alpha}_{0})\I+\bar{\alpha}_{0}\C_{\x_{0}})((1-\bar{\alpha}_{T})\I+\bar{\alpha}_{T}\C_{\x_{0}})^{-1} \bm{\mu}_{T}\\
    &+ (1-\bar{\alpha}_{T}) ((1-\bar{\alpha}_{T})\I+\bar{\alpha}_{T}\C_{\x_{0}})^{-1}\E_{p(\x_{0}|\y)}[\x_{0}]
\end{split}
\end{align}
and $\bm{\mu}_{0}\rightarrow \E_{p(\x_{0}|\y)}[\x_{0}]$ as $\bar{\alpha}_{T}\rightarrow 0$.

\section{Proof of Proposition \ref{prop_1}}\label{proof_prop1}
We have
\begin{align}
\begin{split}
    p(\x_{k}|\x_{k+1:T},\y) &= \frac{p(\x_{k:T},\y)}{p(\x_{k+1:T},\y)}\\
    &=\frac{\int p(\x_{0})p(\y|\x_{0})p(\x_{k:T}|\x_{0})d\x_{0}}{\int p(\x_{0})p(\y|\x_{0})p(\x_{k+1:T}|\x_{0})d\x_{0}}\\
    &=\frac{p(\x_{k+1:T}|\x_{k})\int p(\x_{0})p(\y|\x_{0})p(\x_{k}|\x_{0})d\x_{0}}{p(\x_{k+2:T}|\x_{k+1})\int p(\x_{0})p(\y|\x_{0})p(\x_{k+1}|\x_{0})d\x_{0}}\\
    &=\frac{p(\x_{k+1}|\x_{k})\int p(\x_{0})p(\y|\x_{0})p(\x_{k}|\x_{0})d\x_{0}}{\int p(\x_{0})p(\y|\x_{0})p(\x_{k+1}|\x_{0})d\x_{0}}\\
    &=\frac{p(\x_{k},\x_{k+1},\y)}{p(\x_{k+1},\y)} = p_{k}(\x_{k}|\x_{k+1},\y).
\end{split}
\end{align}

We minimize the KL divergence between variational joint posterior $q(\x_{0:T}|\y) = q(\x_T|\y)\prod_{k=T-1}^{0}q_k(\x_k|\x_{k+1},\y)$ and joint posterior $p(\x_{0:T}|\y) = p(\x_T|\y)\prod_{k=T-1}^{0} p(\x_{k}|\x_{k+1:T},\y)$ to obtain the optimal variational distribution $q$:
\begin{align}
\begin{split}
    \mathcal{F} &=\int q(\x_{0:T}|\y) \log \frac{q(\x_{0:T}|\y)}{p(\x_{0:T}|\y)}d \x_{0:T}\\
    &= \sum_{k=T-1}^{0}\int q(\x_{0:T}|\y)\log \frac{q_k(\x_k|\x_{k+1},\y)}{ p(\x_{k}|\x_{k+1:T},\y) }  d \x_{0:T}\\
    &=\sum_{k=T-1}^{0}\int q(\x_{k:T}|\y)\log \frac{q_k(\x_k|\x_{k+1},\y)}{ p(\x_{k}|\x_{k+1:T},\y) }  d \x_{k:T}\\
    &=\sum_{k=T-1}^{0}\int q(\x_{k:T}|\y)\log \frac{q_k(\x_k|\x_{k+1},\y)}{ p_{k}(\x_{k}|\x_{k+1},\y) }  d \x_{k:T}\\
    &=\sum_{k=T-1}^{0}\int q(\x_{k:k+1}|\y)\log \frac{q_k(\x_k|\x_{k+1},\y)}{ p_{k}(\x_{k}|\x_{k+1},\y) }  d \x_{k}d \x_{k+1}\\
    &=\sum_{k=T-1}^{0}\int q(\x_{k+1}|\y) \int q_{k}(\x_{k}|\x_{k+1},\y) \log \frac{q_k(\x_k|\x_{k+1},\y)}{ p_{k}(\x_{k}|\x_{k+1},\y) }  d \x_{k}d \x_{k+1}
\end{split}
\end{align}
where the last line follows from
\begin{align}
\begin{split}
    q(\x_{k:k+1}|\y) &= \int q(\x_T|\y)\prod_{i=k}^{T-1} q_i(\x_{i}|\x_{i+1},\y)d\x_{k+2:T}\\
    &=\int q_{k}(\x_k|\x_{k+1},\y) q_{k+1}(\x_{k+1}|\x_{k+2},\y)q(\x_{k+2:T}|\y) d\x_{k+2:T}\\
    &= q_{k}(\x_k|\x_{k+1},\y) \int q_{k+1}(\x_{k+1}|\x_{k+2},\y)q(\x_{k+2:T}|\y) d\x_{k+2:T}\\
    &=q_{k}(\x_k|\x_{k+1},\y) q(\x_{k+1}|\y).
\end{split}
\end{align}
Thus, the minimization of $\mathcal{F}$ is equivalent to the minimization of
\begin{align}
    \mathcal{F}_{k} = \int q_{k}(\x_{k}|\x_{k+1},\y)\log \frac{q_k(\x_k|\x_{k+1},\y)}{p_{k}(\x_k|\x_{k+1},\y)} d \x_{k}, \forall k=0,1,\cdots T-1.
\end{align}

\section{Gaussian Mixture Model}\label{gaussian_mixture}
The data prior of $x_0$ is given by
\begin{align}
\begin{split}
    p(x_0) &= \frac{1}{2}\mathcal{N}(x_0;\mu_1,v_1^2)+\frac{1}{2} \mathcal{N}(x_0;\mu_2,v_2^2)\\
    &= \frac{1}{2} (p_1(x_0) +p_2(x_0)).
\end{split}
\end{align}
Then, the data score is 
\begin{align}
    \nabla_{x_0}\log p(x_0) = \frac{1}{2 p(x_0)}\left(-p_1(x_0)\frac{x_0-\mu_1}{\sigma_1^2}-p_2(x_0)\frac{x_0-\mu_2}{\sigma_2^2}\right).
\end{align}

The measurement $y = a x_0 + \sigma_0 \varepsilon_1$ and for VP diffusion, $\x_k = \sqrt{\bar{\alpha}_k}x_0+\sqrt{1-\bar{\alpha}_k}\varepsilon_2$ where $\varepsilon_1$ and $\varepsilon_2$ are i.i.d. from $\mathcal{N}(0,1)$. Then,
\begin{align}
    x_0 = \frac{x_k -  \sqrt{1-\bar{\alpha}_k} \varepsilon_2}{\sqrt{\bar{\alpha}_k}}
\end{align}
and 
\begin{align}
    y = a \frac{x_k - \sqrt{1-\bar{\alpha}_k} \varepsilon_2}{\sqrt{\bar{\alpha}_k}}+ \sigma_0 \varepsilon_1.
\end{align}
Thus, $p(y|x_k) \sim \mathcal{N}(\frac{a x_k}{\sqrt{\bar{\alpha}_k}},\frac{a^2(1-\bar{\alpha}_k)}{\bar{\alpha}_k}+\sigma_0^2)$ and the likelihood score 
\begin{align}
    \nabla_{x_k}\log p(y|x_k) = \frac{a}{\sqrt{\bar{\alpha}_k}}\frac{y-\frac{a}{\sqrt{\bar{\alpha}_k}} x_k}{\frac{a^2(1-\bar{\alpha}_k)}{\bar{\alpha}_k}+\sigma_0^2}.
\end{align}

\section{RMP Implementation Details}\label{params_setting}
\subsection{Parameters setting}
In our experiments, the step size $s_1$ and $\zeta$ for different tasks on FFHQ dataset for VP/VE-RMP are set according to Table \ref{vprmp_parameters} and Table \ref{vermp_parameters} respectively.

\begin{table}[!ht]
\caption{Hyper-parameters of VP-RMP for different tasks}
\footnotesize
\setlength\tabcolsep{3pt}
\renewcommand{\arraystretch}{1.5}
\centering
\begin{tabular}{l|l|l|l|l|l|l|l}
\hline
parameter & SR ($4 \times$)   & box & random & Gauss & motion & nonlinear & phase \\ \hline
$s_1$ & 0.9 & 0.6   &   0.6   & 0.9     &  0.9  & 0.8& 0.5            \\  
$\zeta$       & 0.15  & 0.3          &   0.3           &   0.5           &  0.5    & 0.8&   0.5        \\ \hline
\end{tabular}\label{vprmp_parameters}
\end{table}

\begin{table}[!ht]
\caption{Hyper-parameters of VE-RMP for different tasks}
\centering
\footnotesize
\setlength\tabcolsep{3pt}
\renewcommand{\arraystretch}{1.5}
\begin{tabular}{l|l|l|l|l|l|l|l}
\hline
parameter & SR ($4 \times$)   & box & random & Gauss & motion & nonlinear & phase \\ \hline
$s_1$ & 0.1 & 0.05          &   0.1             & 0.05               &  0.05      & 0.05  &   0.05      \\  
$\zeta$     & 0.07  & 0.1          &   0.25          &   0.15           &    0.15      & 0.15 &  0.35     \\ \hline
\end{tabular}\label{vermp_parameters}
\end{table}

\subsection{Running Time Comparisons}
\begin{figure}
    \centering
    \includegraphics[width=0.8\textwidth]{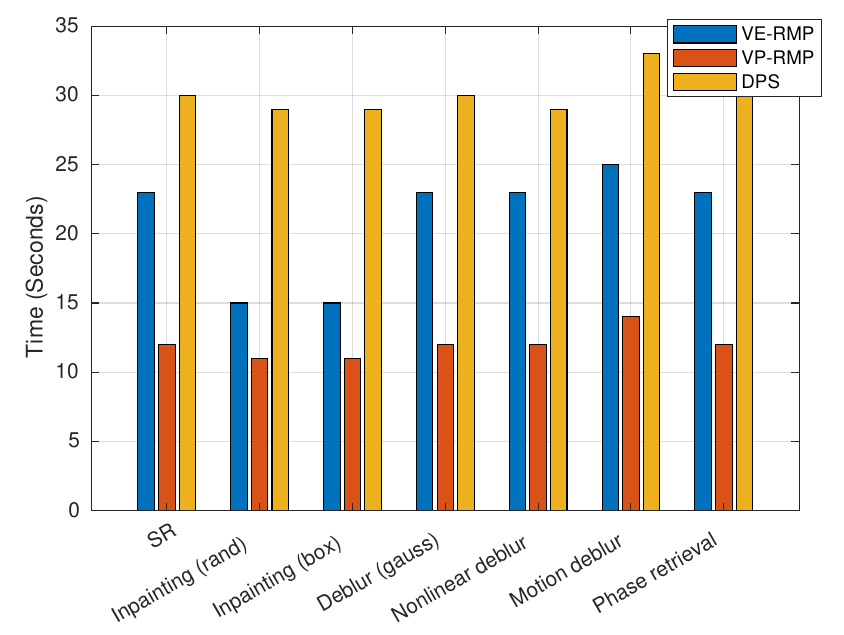}
    \caption{Running time of RMP and DPS for different image reconstruction tasks.}
    \label{fig:running_time}
\end{figure}

On FFHQ dataset, the running time of RMP and DPS for different image reconstruction tasks are shown in Fig. \ref{fig:running_time}. From Fig. \ref{fig:running_time}, we see that VE/VP-RMP achieves a better performance with less running time for all tasks.

\subsection{Approximation of Likelihood Score for Natural Images}
In addition to the approximation of likelihood score we used in the main results for image reconstruction tasks, other approximation methods \cite{meng2022diffusion, song2023pseudoinverse} of likelihood score can be employed to RMP. We present the performance of VP-RMP with the approximation method used in $\Pi$GDM in Table \ref{table_likeli_approximation}. It is shown that the performance of VP-RMP with likelihood approximation method in \cite{song2023pseudoinverse} achieves similar performance as that of VP-RMP with DPS approximation.

\begin{table}[!ht]
\centering
\caption{The performance of VP-RMPs with different approximations of likelihood score on FFHQ dataset.}
\footnotesize
\setlength\tabcolsep{3pt}
\renewcommand{\arraystretch}{1.2}
\begin{tabular}{ccccc}
\hline
\multirow{2}{*}{Methods}       & \multicolumn{4}{c}{SR (4 $\times$)}                                            \\ \cline{2-5} 
     & \multicolumn{1}{c}{PSNR } & SSIM          & \multicolumn{1}{c}{FID } & LPIPS                                \\ \hline
VP-RMP (with approximation in (\ref{likelihood_score_approx})) & ${28.86}$     & ${0.8413}$ &  ${59.7112}$   & ${0.1825}$ \\ 
VP-RMP (with approximation in $\Pi$GDM \cite{song2023pseudoinverse})            &  ${28.70}$     &    ${0.8477}$     & ${61.8101}$  &    ${0.1988}$     \\ 
\hline
\end{tabular}\label{table_likeli_approximation}
\end{table}

\section{Image Reconstruction Results}\label{results}
\begin{figure}[!ht]
    \centering
    \includegraphics[width=0.81\textwidth]{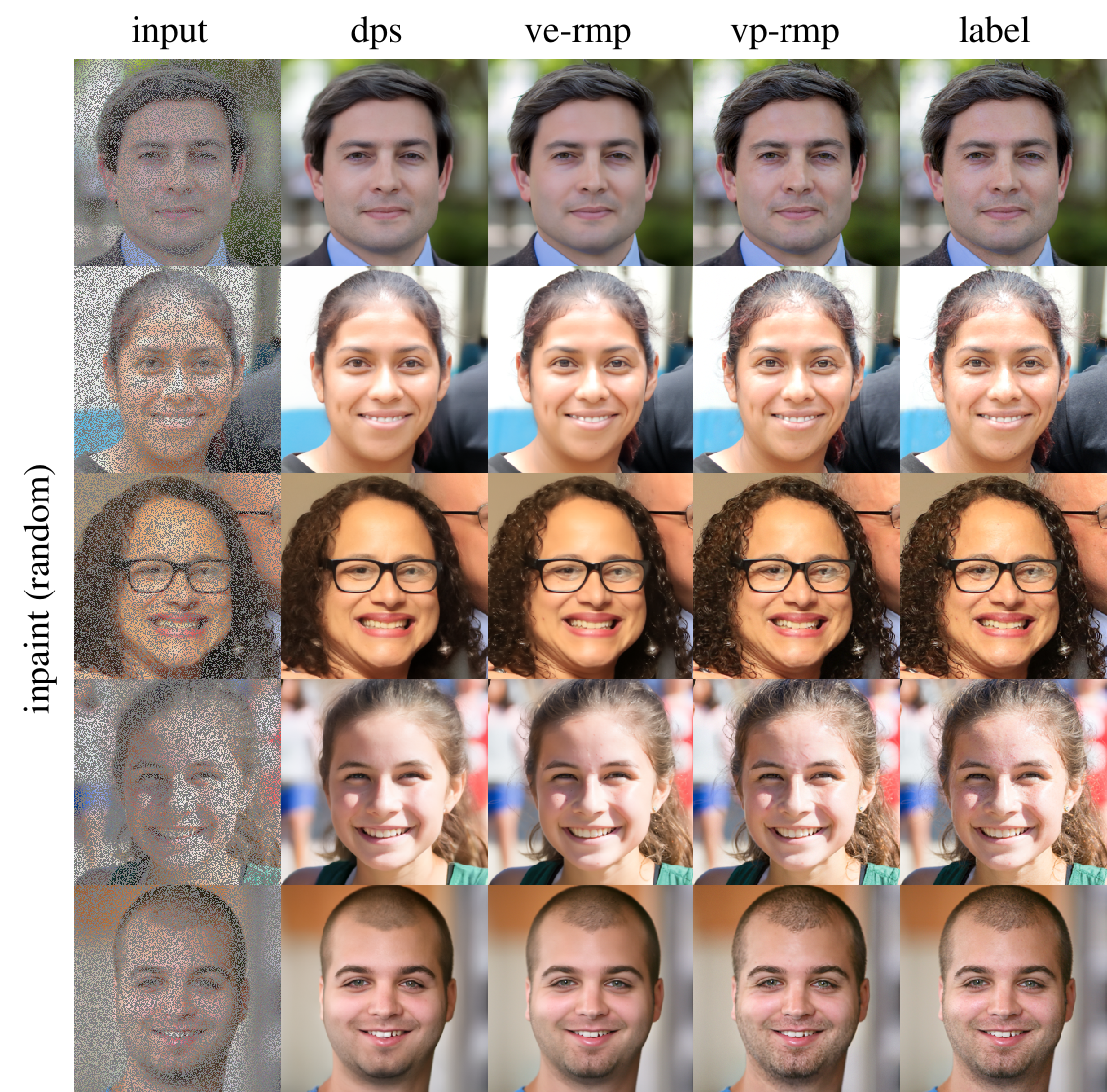}
    \includegraphics[width=0.81\textwidth]{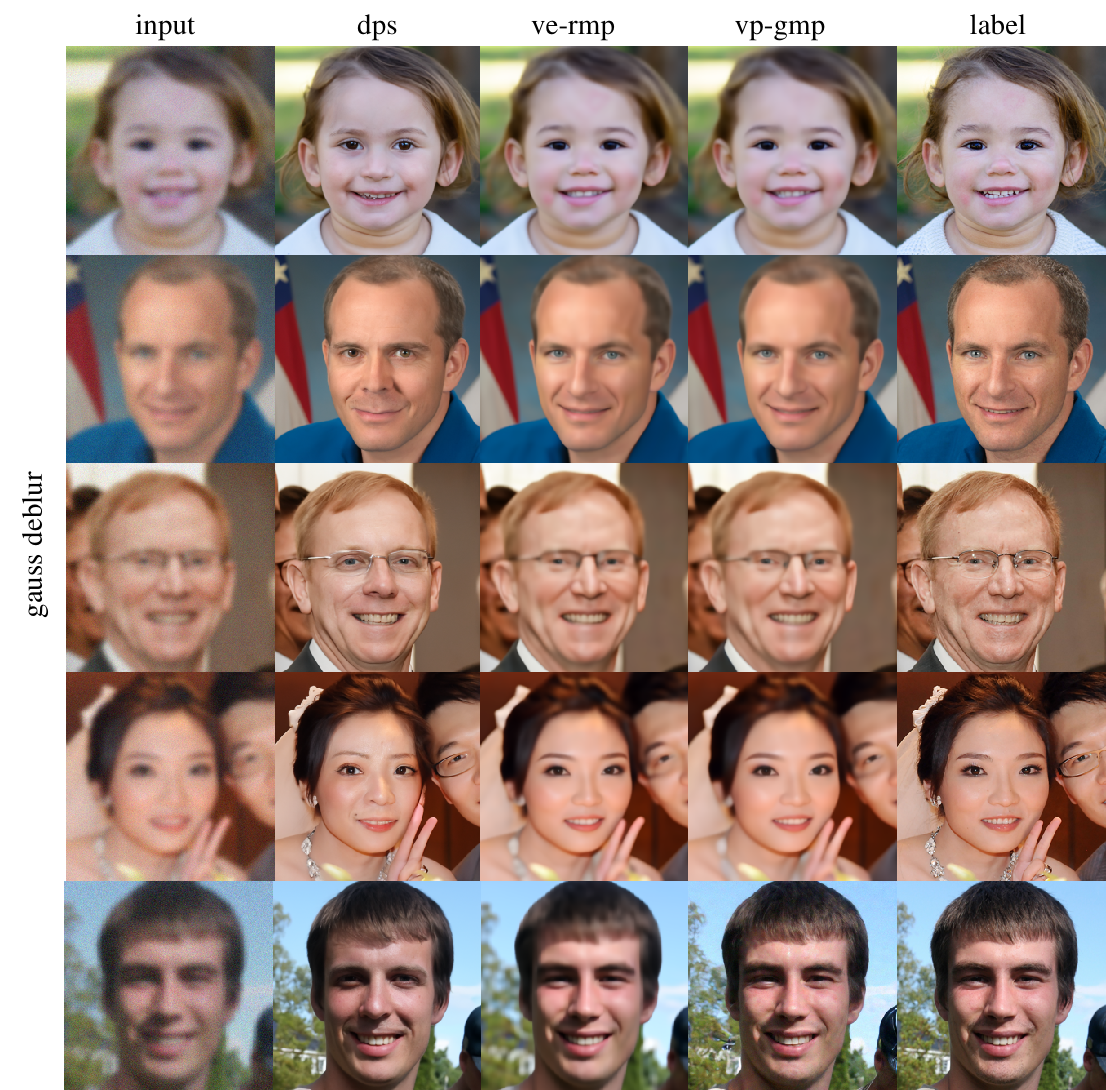}
    \caption{Inpainting (Random) and Gaussian deblur ($\epsilon=0.05$).}
    \label{fig:inpaint_rand_gauss}
\end{figure}

\begin{figure}[!ht]
    \centering
    \includegraphics[width=0.6\textwidth]{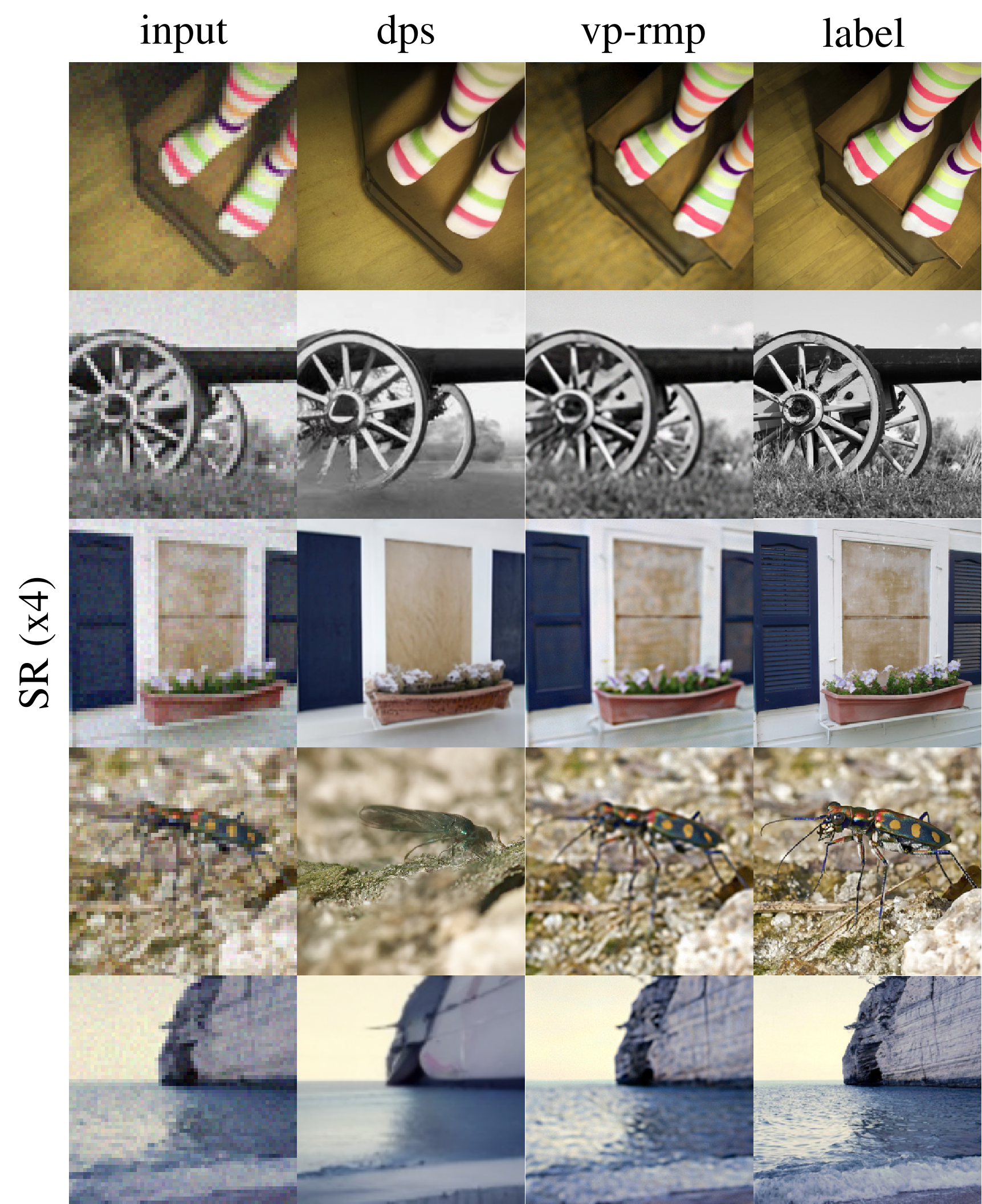}
    \includegraphics[width=0.6\textwidth]{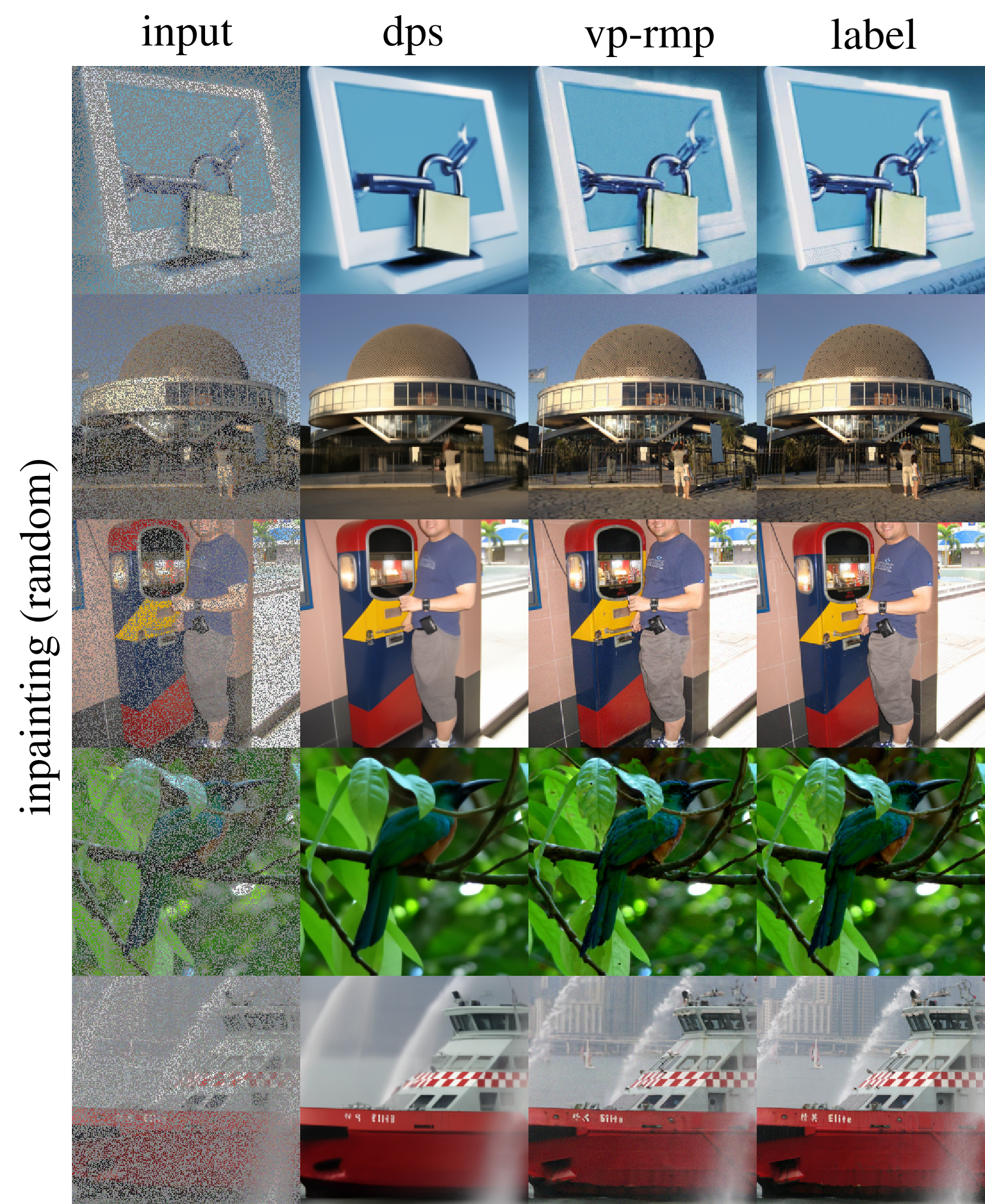}
    \caption{SR ($\times 4$) and Gaussian deblur ($\epsilon=0.05$).}
    \label{fig:image_net_sr_inpaint}
\end{figure}

The image reconstruction results for inpainting (random) and Gaussian deblur tasks on FFHQ dataset are shown in Fig. \ref{fig:inpaint_rand_gauss} and more results of various image reconstruction tasks on ImageNet are presented in Fig. \ref{fig:image_net_sr_inpaint}.

\begin{table}[!ht]
\centering
\caption{Quantitative evaluation of algorithms on ImageNet dataset.}
\footnotesize
\setlength\tabcolsep{2pt}
\renewcommand{\arraystretch}{1.2}
\begin{tabular}{ccccccccccc}
\hline
\multirow{2}{*}{Methods}       & \multicolumn{2}{c}{SR (4 $\times$)}                       & \multicolumn{2}{c}{Inpaint (box)}               & \multicolumn{2}{l}{Inpaint (random)} & \multicolumn{2}{c}{Deblur (Gauss) }                & \multicolumn{2}{c}{Deblur (motion) }                \\ \cline{2-11} 
     & \multicolumn{1}{c}{PSNR $\uparrow$} & SSIM $\uparrow$         & \multicolumn{1}{c}{PSNR $\uparrow$} & SSIM $\uparrow$                 & \multicolumn{1}{l}{PSNR $\uparrow$}    & SSIM $\uparrow$  & \multicolumn{1}{c}{PSNR $\uparrow$} & SSIM $\uparrow$                 & \multicolumn{1}{c}{PSNR $\uparrow$} & SSIM $\uparrow$                 \\ \hline
VP-RMP (Ours) & $\bm{26.21}$     & $\bm{0.7261}$ &  $\bm{19.29}$   & $\bm{0.7830}$  &   $\bm{29.71}$     &  ${0.8720}$       & $\bm{24.65}$    & ${0.6806}$ &   $\bm{25.30}$  & $\bm{0.6533}$ \\ 
DPS        & \multicolumn{1}{c}{21.24}     &  0.5655       & 17.96    &                0.7274       & 27.11     &  0.7862       & 22.49    &  $0.5996$     &   19.72   &  ${0.5118}$                      \\ 
DDRM            &  ${24.28}$     &    ${0.7137}$     & ${19.01}$  &    ${0.7801}$          & ${29.01}$       & $\bm{0.8742} $       & 23.81    &   $\bm{0.7109}$                   & \multicolumn{1}{c}{-}     &   -  \\
 \hline
\multirow{2}{*}{Methods}       & \multicolumn{2}{c}{SR (4 $\times$)}                       & \multicolumn{2}{c}{Inpaint (box)}               & \multicolumn{2}{l}{Inpaint (random)} & \multicolumn{2}{c}{Deblur(Gauss) }                & \multicolumn{2}{c}{Deblur (motion) }                \\ \cline{2-11} 
     & \multicolumn{1}{c}{FID $\downarrow$} & LPIPS $\downarrow$         & \multicolumn{1}{c}{FID $\downarrow$} & LPIPS $\downarrow$                 & \multicolumn{1}{l}{FID $\downarrow$}    & LPIPS $\downarrow$  & \multicolumn{1}{c}{FID $\downarrow$} & LPIPS $\downarrow$                 & \multicolumn{1}{c}{FID $\downarrow$} & LPIPS $\downarrow$                 \\ \hline
VP-RMP (Ours)   & $\bm{75.76}$     &   $\bm{0.3021}$    &   $\bm{106.73}$    &  $\bm{0.2156}$    &   $\bm{26.45}$     &  ${0.1400}$       &  $\bm{85.92}$   & ${0.3105}$ &  
$\bm{71.00}$    & $\bm{0.2933}$ \\
DPS     &${187.14}$     &    ${0.4147}$  & ${146.54}$     &  ${0.2664}$                      & 76.20       &    0.2424     & ${86.28}$    & ${0.3224}$           & ${144.43}$      &   ${0.3795 }$  \\ 
DDRM             &${117.82}$     &        ${0.3086}$      &    ${119.65}$ &     ${0.2232}$        & ${27.18}$        &   $\bm{0.1066}$      & ${95.07}$    &     $\bm{0.2931 }$                 & \multicolumn{1}{c}{-}     &   -  \\\hline
\end{tabular}\label{table_psnr_ssim2}
\end{table}

\end{document}